\documentclass{article}
\usepackage{multirow}
\usepackage[preprint]{neurips_2026}
\usepackage{natbib}
\usepackage{amsmath}
\usepackage{amssymb}
\usepackage{pifont}
\usepackage{microtype}
\usepackage{framed}
\usepackage{wrapfig}
\usepackage{mdframed}
\usepackage{csquotes}
\usepackage{hyperref}
\usepackage{tikz}
\usepackage{algorithm}
\usepackage{algorithmic}
\usepackage{subfig}
\usepackage{newfloat}
\usepackage{listings}
\usepackage{xspace} 
\usepackage{amsfonts}
\usepackage{amsthm}
\usepackage{microtype}
\usepackage{graphicx}
\usepackage{enumitem}
\usepackage{booktabs}
\usepackage{csquotes}
\usepackage{pifont}
\usepackage{makecell}
\usepackage{booktabs} 
\usepackage{algorithm}
\usepackage{graphicx}
\usepackage{subcaption} 
\usepackage{appendix} 
\usepackage{subfig}

\usepackage[utf8]{inputenc} 
\usepackage[T1]{fontenc}    
\usepackage{hyperref}       
\usepackage{url}      
\usepackage{csquotes}
\usepackage{booktabs}       
\usepackage{amsfonts}       
\usepackage{nicefrac}       
\usepackage{microtype}      
\usepackage{xcolor}         
\usepackage{graphicx}
\usepackage{soul} 

\definecolor{bgblue}{rgb}{0.85,0.92,1.0}   
\definecolor{bggreen}{rgb}{0.88,1.0,0.88}  
\definecolor{bgpink}{rgb}{1.0,0.85,0.93}   
\definecolor{bgpurple}{rgb}{0.80,0.85,1.0} 

\definecolor{textbrown}{rgb}{0.65,0.16,0.16} 
\definecolor{textblue}{rgb}{0.1,0.3,0.7}     
\definecolor{textpurple}{rgb}{0.5,0.2,0.5}   
\definecolor{textgreen}{rgb}{0.0,0.5,0.2}    

\usepackage{booktabs}
\usepackage{multirow}
\usepackage{makecell}
\usepackage{colortbl}
\usepackage{xcolor}
\usepackage{array}
\usepackage{rotating}      
\usepackage{siunitx}       
\usepackage{wrapfig}
\usepackage{listings}
\usepackage{pifont}
\definecolor{trtrcolor}{RGB}{214,227,245}   
\definecolor{winbg}{RGB}{180,230,180}        
\definecolor{winsmallbg}{RGB}{220,245,220}   
\definecolor{tiebg}{RGB}{255,248,195}        
\definecolor{lossbg}{RGB}{255,205,205}       
\definecolor{wintext}{RGB}{0,110,0}
\definecolor{losstext}{RGB}{170,0,0}
\definecolor{tietext}{RGB}{130,100,0}

\newcommand{\dwin}[1]{\textcolor{wintext}{\textbf{#1}}}
\newcommand{\dloss}[1]{\textcolor{losstext}{\textit{#1}}}
\newcommand{\dtie}[1]{\textcolor{tietext}{#1}}

\newcolumntype{C}[1]{>{\centering\arraybackslash}p{#1}}

\newcolumntype{C}[1]{>{\centering\arraybackslash}p{#1}}
\newcolumntype{R}[1]{>{\raggedleft\arraybackslash}p{#1}}

\title{\textsc{ReGeN}: Reference-Guided Synthetic Multivariate Time Series Generation for Forecasting}

\author{
Moulik Gupta\textsuperscript{1},
Dhruv Kumar\textsuperscript{1,2},
Murari Mandal\textsuperscript{1,3,$\dagger$},
Saurabh Deshpande\textsuperscript{1,$\dagger$} \\
\\
\textsuperscript{1} Birla AI Labs, Office of Ananya Birla \\
\textsuperscript{2} Birla Institute of Technology and Science, Pilani \\
\textsuperscript{3} Kalinga Institute of Industrial Technology, Bhubaneswar \\
\textsuperscript{$\dagger$} Equal Supervision \\
\texttt{\{moulik.gupta-c, dhruv.kumar-c, murari.mandal-c,} \\
\texttt{saurabh.deshpande-c\}@oab.adityabirla.com}
}

\begin{document}
\maketitle

\begin{abstract}
Training robust multivariate time series forecasting models requires large, diverse corpora, yet many real-world domains provide only a handful of observed sequences. Existing generators fail to resolve this mismatch: prior-based approaches (e.g., CauKer, TimePFN) produce domain-agnostic samples, while data-driven methods (e.g., TimeGAN) treat references as black-box supervision, forfeiting explicit control over periodic structure, local variability, and cross-variable dynamics. We propose \textsc{ReGeN}, a reference-guided generative pipeline that treats observed sequences not as examples to imitate, but as structural scaffolds for controllable synthesis. \textsc{ReGeN} decomposes each reference into three interpretable components: a phase-aligned periodic backbone capturing dominant domain morphology; per-variable stochastic residuals modeled with a deep-kernel Gaussian process; and lag-aware cross-variable dependencies injected through a structural causal model with fitted coupling coefficients. Sampling these components at controllable temperature broadens distributional coverage while preserving domain-grounded structure. We show that \textsc{ReGeN}-generated data consistently substitutes for real sibling data with minimal forecasting degradation, and in strongly periodic domains such as traffic, can outperform the real source itself. We further show that a foundation model pretrained on \textsc{ReGeN} corpora outperforms those pretrained on prior-based and data-driven synthetic alternatives. This suggests that in low-data regimes, how reference data is structurally exploited can matter as much as how much data is available.

\end{abstract}


\section{Introduction}
\label{sec:introduction}

Multivariate time series forecasting underpins decision-making across 
energy grids, traffic networks, cloud infrastructure, and climate systems. Yet their practical deployment remains bottlenecked by data scarcity: most operational multivariate corpora, from building energy 
portfolios to regional sensor networks to clinical monitoring deployments, contain tens to low hundreds of observed sequences per domain, far too few to train robust forecasting models that generalize beyond the observed distribution.~\citep{liu2024itransformer,zeng2023dlinear,wang2025smamba,chronos,moirai}. Fine-tuning a pretrained model can partially compensate, but only when the target domain is already represented in the pretraining corpus, which is rarely the case for niche industrial, environmental, or infrastructure settings. What is needed is a generation strategy that can read domain-specific structure from a small reference corpus and write it into a larger synthetic one.

Existing synthetic generators fall into two categories, each with a fundamental limitation. \textbf{Prior-based generators} (e.g., ForecastPFN~\citep{dooley2023forecastpfn}, TimePFN~\cite{taga2025timepfn}, CauKer~\citep{xie2025cauker}) construct synthetic data from domain-agnostic mathematical primitives (sinusoidal templates, Gaussian Process kernel banks, randomly sampled DAGs) with no grounding in any target domain. The resulting corpora may be statistically plausible but carry none of the domain-specific morphology, uncertainty texture, or cross-variable coupling that a forecasting model must actually learn. \textbf{Data-driven generators} (e.g., TimeGAN~\citep{yoon2019timegan}, C-RNN-GAN~\cite{mogren2016c}) learn directly from observed data, but treat the entire reference corpus as a black-box training signal for a generative model. This approach requires large collections of real sequences to train reliably, and produces generators with no explicit control over periodic structure, local variability, or cross-variable dependencies. Neither approach is well-suited to the practitioner who has access to a small-to-moderate number of real multivariate sequences from a target domain.

We argue that the gap between these two regimes stems from a missed 
opportunity: the observed sequences themselves encode rich, 
domain-specific structure that neither mathematical priors nor 
black-box generators can recover. Even a handful of real sequences 
from an energy building encodes its characteristic demand cycle, 
peak-load uncertainty, and directed coupling between temperature 
and cooling load. This structure is not a limitation to work 
around; it is a signal to exploit. A generation strategy that fits 
every component to observed data would complement both camps: it 
would not require large corpora, nor sacrifice domain grounding, 
addressing the regime where most practitioners actually operate.
 
This motivates a reference-guided approach to synthetic generation in which every component of the generative pipeline is fitted to observed data from the target domain, rather than sampled from generic priors or learned from a black-box objective. We propose \textsc{ReGEN}, a pipeline that treats observed sequences as a structural scaffold for synthesis by decomposing each reference into three interpretable layers: a \textbf{phase-aligned periodic template} to capture the dominant rhythmic backbone of the domain; \textbf{per-variable stochastic residuals} modeled by a deep-kernel Gaussian process to reproduce local uncertainty around that backbone; 
and a \textbf{graph-based lag coupling structure} inferred from the reference data to encode directed cross-variable dependencies. New synthetic sequences are generated by composing samples from each layer at controllable temperature, broadening distributional coverage while preserving the structural character of the target domain.

\textbf{Our key contributions are as follows.}
\begin{itemize}[leftmargin=1.5em,itemsep=2pt,topsep=2pt]
 
\item Reference-guided generation pipeline: We introduce 
\textsc{ReGEN}, a modular generator that grounds all three synthesis 
components in real domain observations rather than generic priors. 
Component ablations confirm that each layer contributes measurably 
to downstream performance.
 
\item Comprehensive empirical validation: We evaluate across twelve datasets spanning five domains, three evaluation protocols (TRTR, TSTR, TRSTR), and five forecasting architectures including a foundation model (Moirai-small). In two-thirds of transfer settings, \textsc{ReGEN} synthetic data substitutes for real sibling data within a $\pm$3\% MSE margin; in strongly periodic domains it outperforms real-data transfer entirely. Training on the union of real and synthetic data yields consistent gains over real-only training for attention- and state-space-based architectures.

\item Superiority over existing generators: We benchmark against both a reference-guided adversarial generator (TimeGAN) and a prior-based causal generator (CauKer) under matched corpus sizes. Foundation models pretrained on \textsc{ReGEN} corpora reduce Moirai MSE by 41\% over TimeGAN and by 2.3\% over CauKer, establishing that \emph{how} reference data is exploited matters as much as \emph{whether} it is available.
\end{itemize}

\section{Related Work}
\label{sec:related}

\textbf{Prior-based and data-driven synthetic generators.} Synthetic time series generation has been pursued along two distinct tracks. Data-driven generators such as TimeGAN~\citep{yoon2019timegan}, C-RNN-GAN~\citep{mogren2016c}, and RCGAN~\citep{esteban2017real} learn directly from a target corpus using adversarial and recurrent objectives. They produce realistic sequences when sufficient training data is available, but operate as black-box models with no explicit control over periodic structure, local variability, or cross-variable coupling, and they require enough real sequences to train the generator itself. Prior-based generators take the opposite approach, constructing synthetic data from domain-agnostic mathematical primitives without requiring any real target data. ForecastPFN~\citep{dooley2023forecastpfn} trains on Bayesian forecasting priors; Chronos augments pretraining with KernelSynth, which composes GP kernels to generate univariate series~\citep{chronos}; TimePFN extends this to multivariate settings via GP kernel composition with linear model of coregionalization~\citep{taga2025timepfn}; CauKer combines GP kernels with a randomly sampled causal DAG to produce causally coherent multivariate sequences~\citep{xie2025cauker}; and SarSim uses SARIMA-based simulation with multi-seasonality and heavy-tailed perturbations for large-scale pretraining~\citep{oreshkin2026sarsim0}. These methods are well-suited to foundation model pretraining but their synthetic structure bears no relationship to any specific target domain, making them unsuitable for dataset-conditioned augmentation.


\textbf{Structural decomposition and augmentation.} A complementary thread draws on classical time series decomposition. STL-style 
seasonal-trend decomposition separates a signal into interpretable components~\citep{cleveland1990stl}, and structured probabilistic models such as Gaussian processes with compositional kernels~\citep{roberts2013gaussian} provide principled uncertainty over residuals. Lightweight augmentation methods such as TSMix~\citep{darlow2023tsmix} and mixup-style 
variants~\citep{aggarwal2023mixuppp} improve downstream performance 
through interpolation, but do not model multivariate dependence or 
directed lag structure. \textsc{ReGEN} occupies a distinct position: 
unlike prior-based generators it conditions all three components on 
real reference observations, and unlike data-driven generators it 
requires only a moderate corpus while providing explicit control over 
periodic structure, residual uncertainty, and cross-variable coupling.
\section{\textsc{ReGen}}
\label{sec:method}


We consider the problem of generating synthetic multivariate time 
series that preserve three salient properties of real temporal 
systems: recurring within-variable structure, stochastic local 
variability, and directed dependence across variables. Let 
$\mathcal{D} = \{\mathbf{X}^{(s)}\}_{s=1}^{S}$ denote a collection 
of real multivariate sequences with 
$\mathbf{X}^{(s)} \in \mathbb{R}^{C \times T}$, where $C$ is the 
number of covariates. Our goal is to construct a generative mechanism 
for synthetic trajectories 
$\widetilde{\mathbf{X}} \in \mathbb{R}^{C \times T_{\mathrm{gen}}}$ 
that reproduces marginal temporal morphology while remaining faithful 
to cross-variable dynamics. Rather than modeling these three sources 
of structure monolithically, we explicitly separate (i) a 
low-frequency periodic backbone, (ii) a stochastic innovation 
process, and (iii) a graph-structured interaction mechanism, so that 
periodicity, uncertainty, and causal dependence can be modeled and 
controlled independently.

\begin{figure}[t]
\centering
\includegraphics[width=0.99\textwidth]{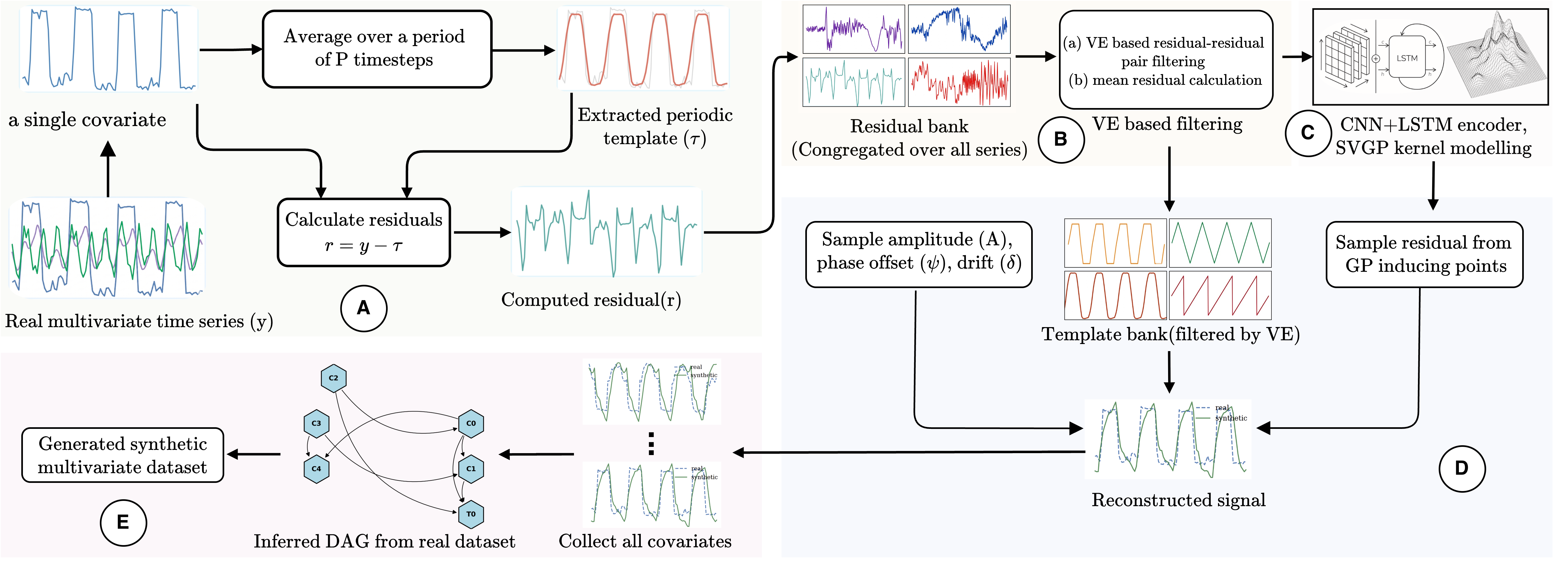}
\caption{\textsc{ReGeN} pipeline overview. \textbf{A:} Extract a phase-aligned periodic template and compute residuals from the real multivariate time series. \textbf{B:} Aggregate residuals across series and apply VE-based filtering to retain reliable template--residual structure. \textbf{C:} Fit a CNN+LSTM encoder with an SVGP-based deep kernel prior to model residual dynamics. \textbf{D:} Sample template parameters and GP residuals, then combine them to reconstruct synthetic signals. \textbf{E:} Use the inferred DAG to inject cross-variate dependencies and assemble the final synthetic multivariate dataset.}
\label{fig:architecture}
\end{figure}

\textbf{Structural decomposition of each covariate.} For each covariate \(c\), we decompose the observed trajectory in normalized space as
\begin{equation}
 y_{c,t} = \tau_{c,t} + r_{c,t},
\end{equation}
where \(\tau_{c,t}\) denotes a recurring structural template and \(r_{c,t}\) denotes the residual component. The template is intended to capture phase-locked regularity, while the residual absorbs local departures from this repeating pattern.
This separation between recurring structure and stochastic remainder is directly inspired by the classical seasonal decomposition and structured probabilistic modelling of temporal signals \citep{cleveland1990stl,roberts2013gaussian}. To estimate \(\tau_{c,t}\), we align observations according to a characteristic period \(P\) (specified dataset-wise in Appendix~\ref{app:dataset_details}, Table~\ref{tab:dataset_summary}) and average values sharing the same phase. For a real sequence indexed by \(s\), this yields a phase template
\begin{equation}
\tau^{(s)}_c(p) = \frac{1}{|\mathcal{I}_p|} \sum_{t \in \mathcal{I}_p} y^{(s)}_{c,t},
\qquad
\mathcal{I}_p = \{t \mid t \bmod P = p\},
\end{equation}
Here, \(p \in \{0,\dots,P-1\}\) indexes the phase within a period of length \(P\), \(t\) indexes discrete time steps, \(y^{(s)}_{c,t}\) is the normalized observation of covariate \(c\) at time \(t\) in sample \(s\), \(\mathcal{I}_p\) is the set of all time indices assigned to phase \(p\). The residual is then defined as
\begin{equation}
 r^{(s)}_{c,t} = y^{(s)}_{c,t} - \tau^{(s)}_c(t \bmod P).
\end{equation}

This decomposition is motivated by the observation that many real-world temporal systems contain a strong periodic component whose amplitude and phase are relatively stable, even when short-term fluctuations remain highly stochastic. To retain only structurally meaningful templates, we score each decomposition by the fraction of signal variance explained (VE) by the periodic component,
\begin{equation}
\mathrm{VE}^{(s)}_c = 1 - \frac{\mathrm{Var}(\mathbf{r}^{(s)}_c)}{\mathrm{Var}(\mathbf{y}^{(s)}_c)}.
\end{equation}
High-scoring decompositions define a library of representative templates for each covariate, together with empirical amplitude statistics that characterize how strongly the periodic component manifests across real samples. We fit the residual model on a filtered mean signal built from the samples that pass the VE filter,
\begin{equation}
\bar{y}_{c,t} = \frac{1}{|\mathcal{S}_c|} \sum_{s \in \mathcal{S}_c} y^{(s)}_{c,t},
\qquad
\mathcal{S}_c = \{s \mid \mathrm{VE}^{(s)}_c \ge \eta\},
\end{equation}
where \(\eta\) is the VE threshold. We then compute the periodic template of this filtered mean signal using the same phase-averaging procedure,
\begin{equation}
\bar{\tau}_c(p) = \frac{1}{|\mathcal{I}_p|} \sum_{t \in \mathcal{I}_p} \bar{y}_{c,t},
\end{equation}
and define the mean residual as
\begin{equation}
\bar{r}_{c,t} = \bar{y}_{c,t} - \bar{\tau}_c(t \bmod P).
\end{equation}
While averaging suppresses series-specific variability, the resulting residual signal preserves shared temporal dynamics. In this sense, \(\bar{r}_{c,t}\) acts as a cleaned residual that washes away idiosyncratic noise while retaining the common residual structure, making it easier to fit the DKL model.

\textbf{Template-based structural scaffold.} After extracting a periodic template for each covariate, we use it to define the coarse structure of a synthetic multivariate sample. For a selected template \(\tau_{c,m}\), where \(m\) denotes the \(m\)-th template for covariate \(c\), the structural contribution is
\begin{equation}
\widehat{\tau}_{c,t} = \bar{\tau}_{c,m} + a_c \big(\tau_{c,m}(t \bmod P) - \bar{\tau}_{c,m}\big),
\end{equation}
where \(\bar{\tau}_{c,m}\) is the template mean and \(a_c\) is a covariate-specific amplitude factor. This preserves the periodic morphology while allowing different covariates to express it with different strengths.

\textbf{Residual dynamics via deep kernel autoregression.} After constructing the cleaned mean residual, we model its short-range dependence, local roughness, and uncertainty autoregressively using a deep kernel learning construction for sequential data \citep{wilson2016deep,alshedivat2017learning}. For covariate \(c\), let
\begin{equation}
\mathbf{u}_{c,t} = [\bar{r}_{c,t-W}, \dots, \bar{r}_{c,t-1}]
\end{equation}
denote a context window of length \(W\). This context is mapped to a latent representation through a neural feature extractor,
\begin{equation}
\mathbf{z}_{c,t} = g_{\theta}(\mathbf{u}_{c,t}) \in \mathbb{R}^{d},
\end{equation}
where \(g_{\theta}\) is implemented as a convolutional-recurrent encoder with an LSTM recurrent block \citep{hochreiter1997long}. Convolutional layers capture local motifs and short-range temporal patterns, while a recurrent block summarizes their sequential evolution over the window.

The latent state \(\mathbf{z}_{c,t}\) parameterizes a Gaussian process over residual dynamics \citep{rasmussen2006gaussian},
\begin{equation}
 f_c \sim \mathcal{GP}\big(0, k_c(\mathbf{z}, \mathbf{z}')\big),
\qquad
 \bar{r}_{c,t} \mid \mathbf{u}_{c,t} \sim \mathcal{N}(f_c(\mathbf{z}_{c,t}), \sigma_c^2),
\end{equation}
where \(k_c\) is a flexible kernel in latent space. In our case, the kernel combines Matern and radial basis components \citep{rasmussen2006gaussian}, thereby accommodating both rough local behavior and smoother variation. This deep kernel formulation is critical: the neural encoder provides expressive history-dependent features, while the Gaussian process introduces calibrated uncertainty and a principled nonparametric prior over residual trajectories \citep{wilson2016deep,alshedivat2017learning}.

\textbf{Autoregressive synthesis of intrinsic trajectories.} Given the learned residual model, synthetic residuals are generated autoregressively. At each time step, the model outputs a predictive mean and variance conditioned on the previously generated context. Sampling is performed as
\begin{equation}
\widehat{r}_{c,t} = \mu_{c,t} + \lambda_c \sigma_{c,t} \epsilon_t,
\qquad
\epsilon_t \sim \mathcal{N}(0,1),
\end{equation}
where \(\mu_{c,t}\) and \(\sigma_{c,t}\) are the predictive moments of the deep kernel model and \(\lambda_c\) is a temperature parameter controlling stochasticity. Larger temperatures increase diversity, whereas smaller temperatures produce trajectories closer to the posterior mean.

We also add a low-amplitude sinusoidal drift term \(d_{c,t}\) to capture slow variation beyond the template and residual model. The intrinsic synthetic trajectory for covariate \(c\) is then given by
\begin{equation}
\widehat{x}^{\mathrm{int}}_{c,t} = \widehat{\tau}_{c,t} + \widehat{r}_{c,t} + d_{c,t}.
\end{equation}
This intrinsic signal represents the variable-specific dynamics before cross-variable coupling is imposed.

\textbf{Directed coupling across variables.} Real multivariate systems rarely consist of independent channels. To encode directed dependence, we introduce a graph-based interaction layer defined on a directed acyclic graph \(\mathcal{G}\), following the structural causal perspective of DAG-based models and the synthetic time-series generation setting of CauKer \citep{peters2017elements,xie2025cauker}. Each node corresponds to a covariate, and each directed edge specifies a parent-child relationship together with admissible lags. Root variables are generated directly from their intrinsic dynamics, whereas child variables are modified using lagged parent information. Appendix Section~\ref{app:dag_estimation} gives the full consensus causal-discovery procedure used to estimate \(\mathcal{G}\) and its admissible lag sets from the real data.

For a child variable \(c\) with parent set \(\mathrm{Pa}(c)\), we define the aggregated parent contribution as
\begin{equation}
 g_{c,t} = \sum_{p \in \mathrm{Pa}(c)} \sum_{\ell \in \mathcal{L}_{p \to c}} w_{p,c,\ell} \, s_{p,t-\ell},
\end{equation}
where \(\mathcal{L}_{p \to c}\) is the set of admissible lags, \(w_{p,c,\ell}\) are consensus-derived coupling coefficients estimated from the retained discovery scores in Appendix Section~\ref{app:dag_estimation}, and \(s_{p,t}\) denotes the full parent signal. In other words, directed dependence is imposed on the entire generated series, so the structural template and residual dynamics are mixed in a single coupling step.

The final child trajectory is produced through a convex combination of its intrinsic component and the parent-driven term,
\begin{equation}
\widetilde{x}_{c,t} = \alpha_c \widehat{x}^{\mathrm{int}}_{c,t} + (1-\alpha_c) h(g_{c,t}),
\end{equation}
where \(\alpha_c \in [0,1]\) controls the relative strength of endogenous and exogenous dynamics, and \(h\) is optionally a nonlinear transformation. In the implementation, we sample \(\alpha_c\) uniformly from \([0.7, 0.9]\) so that the endogenous structure of each covariate is consistently prioritized over parent-specific dynamics. The nonlinear map serves to model saturating or asymmetric causal effects without destabilizing the generated trajectories.

\providecommand{\rowbest}[1]{\textcolor{textgreen}{#1}}
\providecommand{\rowsecondbest}[1]{#1}
\providecommand{\rowsecond}[1]{#1}
\section{Experiments and Results}
\label{sec:experiments}
We evaluate \textsc{ReGeN} along three complementary axes that together make a cumulative case for reference-guided synthetic generation. 

\textbf{(Q1) Can reference-guided synthetic data substitute for a real sibling dataset?} We test whether a corpus generated from one domain can serve as the sole training source for a model deployed on a closely related domain, replacing the real sibling entirely. 

\textbf{(Q2) Does synthetic augmentation on top of real data push performance beyond what real data alone achieves?} We ask whether \textsc{ReGEN}-generated series add information that the available real corpus does not already contain. 

\textbf{(Q3) Does the \emph{structured decomposition} in \textsc{ReGEN} produce better synthetic data than alternative reference-guided and prior-based generators?} We benchmark against a reference-guided adversarial generator~\cite{yoon2019timegan} and a prior-based causal generator~\cite{xie2025cauker} on exactly matched corpus sizes, isolating generation quality from data volume and reference availability.


Throughout, we emphasize that synthetic data is useful only when it preserves structure that transfers: periodic morphology, residual 
uncertainty, and cross-variable coupling.  Direct structural diagnostics like spectral fidelity, residual calibration, and cross-variate coupling recovery are provided in Appendix \ref{app:ablation}.

\begin{table*}[!t]
\centering
\begingroup
\renewcommand{\dwin}[1]{#1}
\renewcommand{\dtie}[1]{#1}
\renewcommand{\dloss}[1]{#1}
\caption{%
  Zero-shot forecasting transfer across sibling dataset pairs under
  real-to-real (\textbf{TRTR}) and train-on-synthetic, test-on-real
  (\textbf{TSTR}) protocols, evaluated on iTransformer, DLinear, and S-Mamba. Each TSTR row directly below shows REGEN synthetic performance.
  The \textbf{$\Delta$\,(\%)} column reports relative change
  $({\rm TSTR}-{\rm TRTR})/{\rm TRTR}\times 100$ in MSE.
}
\label{tab:compact_all_pairs}
\label{tab:regen_results_3pct}
\setlength{\tabcolsep}{4pt}
\renewcommand{\arraystretch}{1.2}
\resizebox{0.85\textwidth}{!}{%
\begin{tabular}{
    >{\centering\arraybackslash}p{0.7cm}   
    l                                       
    >{\centering\arraybackslash}p{0.8cm}   
    >{\centering\arraybackslash}p{1.0cm}   
    >{\centering\arraybackslash}p{0.6cm}
    >{\centering\arraybackslash}p{0.6cm}
    >{\centering\arraybackslash}p{0.9cm}
    >{\centering\arraybackslash}p{0.6cm}
    >{\centering\arraybackslash}p{0.6cm}
    >{\centering\arraybackslash}p{0.9cm}
    >{\centering\arraybackslash}p{0.6cm}
    >{\centering\arraybackslash}p{0.6cm}
    >{\centering\arraybackslash}p{0.9cm}
  }
\toprule
\multirow{2}{*}{\textbf{Dom.}}
  & \multirow{2}{*}{\textbf{Sibling Pair}}
  & \multirow{2}{*}{\textbf{Type}}
  & \multirow{2}{*}{\textbf{Run}}
  & \multicolumn{3}{c}{\textbf{iTransformer}}
  & \multicolumn{3}{c}{\textbf{DLinear}}
  & \multicolumn{3}{c}{\textbf{S-Mamba}} \\
\cmidrule(lr){5-7}\cmidrule(lr){8-10}\cmidrule(lr){11-13}
  & & & &
  {\small MSE$\downarrow$} & {\small MAE$\downarrow$} & {\small $\Delta$(\%)} &
  {\small MSE$\downarrow$} & {\small MAE$\downarrow$} & {\small $\Delta$(\%)} &
  {\small MSE$\downarrow$} & {\small MAE$\downarrow$} & {\small $\Delta$(\%)} \\
\midrule

\multirow{8}{*}{\rotatebox{90}{\small\textbf{Energy}}}

& \multirow{4}{*}{\makecell[l]{BDG-2 Bear / Panther}}
  & \cellcolor{trtrcolor}TRTR & \cellcolor{trtrcolor}A$\to$B
  & \cellcolor{trtrcolor}0.32 & \cellcolor{trtrcolor}0.36 & \cellcolor{trtrcolor}{---}
  & \cellcolor{trtrcolor}0.29 & \cellcolor{trtrcolor}0.36 & \cellcolor{trtrcolor}{---}
  & \cellcolor{trtrcolor}0.29 & \cellcolor{trtrcolor}0.34 & \cellcolor{trtrcolor}{---} \\
& & TSTR & D$\to$B
  & \cellcolor{lossbg}0.36 & \cellcolor{lossbg}0.38 & \dloss{+12.5}
  & \cellcolor{tiebg}0.30  & \cellcolor{tiebg}0.36  & \dtie{+3.4}
  & \cellcolor{tiebg}0.29  & \cellcolor{tiebg}0.35  & \dtie{0.0} \\[2pt]

&
  & \cellcolor{trtrcolor}TRTR & \cellcolor{trtrcolor}B$\to$A
  & \cellcolor{trtrcolor}0.41 & \cellcolor{trtrcolor}0.40 & \cellcolor{trtrcolor}{---}
  & \cellcolor{trtrcolor}0.41 & \cellcolor{trtrcolor}0.39 & \cellcolor{trtrcolor}{---}
  & \cellcolor{trtrcolor}0.42 & \cellcolor{trtrcolor}0.34 & \cellcolor{trtrcolor}{---} \\
& & TSTR & C$\to$A
  & \cellcolor{tiebg}0.41   & \cellcolor{lossbg}0.43 & \dtie{0.0}
  & \cellcolor{lossbg}0.43  & \cellcolor{lossbg}0.43 & \dloss{+4.9}
  & \cellcolor{winbg}0.40   & \cellcolor{winsmallbg}0.33 & \dwin{$-$4.8} \\[2pt]
\cmidrule(lr){2-13}

& \multirow{4}{*}{\makecell[l]{BDG-2 Bull / Hog}}
  & \cellcolor{trtrcolor}TRTR & \cellcolor{trtrcolor}A$\to$B
  & \cellcolor{trtrcolor}0.50 & \cellcolor{trtrcolor}0.49 & \cellcolor{trtrcolor}{---}
  & \cellcolor{trtrcolor}0.48 & \cellcolor{trtrcolor}0.45 & \cellcolor{trtrcolor}{---}
  & \cellcolor{trtrcolor}0.49 & \cellcolor{trtrcolor}0.46 & \cellcolor{trtrcolor}{---} \\
& & TSTR & D$\to$B
  & \cellcolor{tiebg}0.50   & \cellcolor{winsmallbg}0.48 & \dtie{0.0}
  & \cellcolor{tiebg}0.49   & \cellcolor{tiebg}0.46      & \dtie{+2.1}
  & \cellcolor{winsmallbg}0.48 & \cellcolor{winsmallbg}0.45 & \dtie{$-$2.0} \\[2pt]

&
  & \cellcolor{trtrcolor}TRTR & \cellcolor{trtrcolor}B$\to$A
  & \cellcolor{trtrcolor}0.33 & \cellcolor{trtrcolor}0.40 & \cellcolor{trtrcolor}{---}
  & \cellcolor{trtrcolor}0.36 & \cellcolor{trtrcolor}0.39 & \cellcolor{trtrcolor}{---}
  & \cellcolor{trtrcolor}0.33 & \cellcolor{trtrcolor}0.40 & \cellcolor{trtrcolor}{---} \\
& & TSTR & C$\to$A
  & \cellcolor{lossbg}0.37  & \cellcolor{tiebg}0.40       & \dloss{+12.1}
  & \cellcolor{winbg}0.31   & \cellcolor{winsmallbg}0.38  & \dwin{$-$13.9}
  & \cellcolor{lossbg}0.35  & \cellcolor{tiebg}0.41       & \dloss{+6.1} \\

\midrule

\multirow{4}{*}{\rotatebox{90}{\small\textbf{Cloud}}}

& \multirow{4}{*}{\makecell[l]{Azure VM 2017 / Borg 2011}}
  & \cellcolor{trtrcolor}TRTR & \cellcolor{trtrcolor}A$\to$B
  & \cellcolor{trtrcolor}0.58 & \cellcolor{trtrcolor}0.49 & \cellcolor{trtrcolor}{---}
  & \cellcolor{trtrcolor}0.59 & \cellcolor{trtrcolor}0.54 & \cellcolor{trtrcolor}{---}
  & \cellcolor{trtrcolor}0.58 & \cellcolor{trtrcolor}0.53 & \cellcolor{trtrcolor}{---} \\
& & TSTR & D$\to$B
  & \cellcolor{tiebg}0.59   & \cellcolor{tiebg}0.50  & \dtie{+1.7}
  & \cellcolor{tiebg}0.59   & \cellcolor{winbg}0.52  & \dtie{0.0}
  & \cellcolor{lossbg}0.61  & \cellcolor{lossbg}0.55 & \dloss{+5.2} \\[2pt]

&
  & \cellcolor{trtrcolor}TRTR & \cellcolor{trtrcolor}B$\to$A
  & \cellcolor{trtrcolor}0.88 & \cellcolor{trtrcolor}0.41 & \cellcolor{trtrcolor}{---}
  & \cellcolor{trtrcolor}0.90 & \cellcolor{trtrcolor}0.44 & \cellcolor{trtrcolor}{---}
  & \cellcolor{trtrcolor}0.90 & \cellcolor{trtrcolor}0.41 & \cellcolor{trtrcolor}{---} \\
& & TSTR & C$\to$A
  & \cellcolor{tiebg}0.90   & \cellcolor{lossbg}0.45 & \dtie{+2.3}
  & \cellcolor{tiebg}0.92   & \cellcolor{winbg}0.42  & \dtie{+2.2}
  & \cellcolor{lossbg}0.97  & \cellcolor{tiebg}0.42  & \dloss{+7.8} \\

\midrule

\multirow{4}{*}{\rotatebox{90}{\small\textbf{Traffic}}}

& \multirow{4}{*}{\makecell[l]{PEMS-04 / PEMS-08}}
  & \cellcolor{trtrcolor}TRTR & \cellcolor{trtrcolor}A$\to$B
  & \cellcolor{trtrcolor}0.32 & \cellcolor{trtrcolor}0.34 & \cellcolor{trtrcolor}{---}
  & \cellcolor{trtrcolor}0.30 & \cellcolor{trtrcolor}0.29 & \cellcolor{trtrcolor}{---}
  & \cellcolor{trtrcolor}0.29 & \cellcolor{trtrcolor}0.30 & \cellcolor{trtrcolor}{---} \\
& & TSTR & D$\to$B
  & \cellcolor{winbg}0.30  & \cellcolor{winbg}0.29  & \dwin{$-$6.3}
  & \cellcolor{winbg}0.28  & \cellcolor{winbg}0.28  & \dwin{$-$6.7}
  & \cellcolor{winbg}0.26  & \cellcolor{tiebg}0.30  & \dwin{$-$10.3} \\[2pt]

&
  & \cellcolor{trtrcolor}TRTR & \cellcolor{trtrcolor}B$\to$A
  & \cellcolor{trtrcolor}0.31 & \cellcolor{trtrcolor}0.31 & \cellcolor{trtrcolor}{---}
  & \cellcolor{trtrcolor}0.32 & \cellcolor{trtrcolor}0.31 & \cellcolor{trtrcolor}{---}
  & \cellcolor{trtrcolor}0.30 & \cellcolor{trtrcolor}0.28 & \cellcolor{trtrcolor}{---} \\
& & TSTR & C$\to$A
  & \cellcolor{lossbg}0.37  & \cellcolor{lossbg}0.38 & \dloss{+19.4}
  & \cellcolor{winbg}0.29   & \cellcolor{winbg}0.30  & \dwin{$-$9.4}
  & \cellcolor{winsmallbg}0.29 & \cellcolor{lossbg}0.33 & \dtie{$-$3.3} \\

\midrule

\multirow{4}{*}{\rotatebox{90}{\small\textbf{Climate}}}

& \multirow{4}{*}{\makecell[l]{Subseasonal / Precip.}}
  & \cellcolor{trtrcolor}TRTR & \cellcolor{trtrcolor}A$\to$B
  & \cellcolor{trtrcolor}1.09 & \cellcolor{trtrcolor}0.77 & \cellcolor{trtrcolor}{---}
  & \cellcolor{trtrcolor}0.83 & \cellcolor{trtrcolor}0.60 & \cellcolor{trtrcolor}{---}
  & \cellcolor{trtrcolor}0.76 & \cellcolor{trtrcolor}0.57 & \cellcolor{trtrcolor}{---} \\
& & TSTR & D$\to$B
  & \cellcolor{winbg}1.01   & \cellcolor{winbg}0.73  & \dwin{$-$7.3}
  & \cellcolor{lossbg}0.90  & \cellcolor{lossbg}0.64 & \dloss{+8.4}
  & \cellcolor{winsmallbg}0.74 & \cellcolor{winsmallbg}0.56 & \dtie{$-$2.6} \\[2pt]

&
  & \cellcolor{trtrcolor}TRTR & \cellcolor{trtrcolor}B$\to$A
  & \cellcolor{trtrcolor}0.42 & \cellcolor{trtrcolor}0.46 & \cellcolor{trtrcolor}{---}
  & \cellcolor{trtrcolor}0.40 & \cellcolor{trtrcolor}0.47 & \cellcolor{trtrcolor}{---}
  & \cellcolor{trtrcolor}1.34 & \cellcolor{trtrcolor}1.02 & \cellcolor{trtrcolor}{---} \\
& & TSTR & C$\to$A
  & \cellcolor{winbg}0.40   & \cellcolor{winbg}0.43  & \dwin{$-$4.8}
  & \cellcolor{winbg}0.35   & \cellcolor{winbg}0.39  & \dwin{$-$12.5}
  & \cellcolor{lossbg}1.51  & \cellcolor{lossbg}1.10 & \dloss{+12.7} \\

\midrule

\multirow{4}{*}{\rotatebox{90}{\small\textbf{Energy (res.)}}}

& \multirow{4}{*}{\makecell[l]{Res.\ PV Power / Load Power}}
  & \cellcolor{trtrcolor}TRTR & \cellcolor{trtrcolor}A$\to$B
  & \cellcolor{trtrcolor}0.60 & \cellcolor{trtrcolor}0.42 & \cellcolor{trtrcolor}{---}
  & \cellcolor{trtrcolor}0.55 & \cellcolor{trtrcolor}0.40 & \cellcolor{trtrcolor}{---}
  & \cellcolor{trtrcolor}0.54 & \cellcolor{trtrcolor}0.37 & \cellcolor{trtrcolor}{---} \\
& & TSTR & D$\to$B
  & \cellcolor{winbg}0.54   & \cellcolor{winbg}0.38  & \dwin{$-$10.0}
  & \cellcolor{lossbg}0.64  & \cellcolor{tiebg}0.41  & \dloss{+16.4}
  & \cellcolor{tiebg}0.54   & \cellcolor{tiebg}0.37  & \dtie{0.0} \\[2pt]

&
  & \cellcolor{trtrcolor}TRTR & \cellcolor{trtrcolor}B$\to$A
  & \cellcolor{trtrcolor}0.25 & \cellcolor{trtrcolor}0.22 & \cellcolor{trtrcolor}{---}
  & \cellcolor{trtrcolor}0.25 & \cellcolor{trtrcolor}0.24 & \cellcolor{trtrcolor}{---}
  & \cellcolor{trtrcolor}0.23 & \cellcolor{trtrcolor}0.20 & \cellcolor{trtrcolor}{---} \\
& & TSTR & C$\to$A
  & \cellcolor{lossbg}0.26  & \cellcolor{winbg}0.20  & \dloss{+4.0}
  & \cellcolor{lossbg}0.31  & \cellcolor{lossbg}0.29 & \dloss{+24.0}
  & \cellcolor{winbg}0.22   & \cellcolor{lossbg}0.23 & \dwin{$-$4.3} \\

\bottomrule
\end{tabular}%
}

\smallskip
\noindent
\begin{minipage}{\textwidth}
\footnotesize
\textbf{Colour key (±3\% relative margin):}\quad
\colorbox{trtrcolor}{\strut\ TRTR\ } Real-to-real reference;\quad
\colorbox{winbg}{\strut\ \textcolor{wintext}{\textbf{dark green}}\ } TSTR beats TRTR by ${>}3\%$;\quad
\colorbox{winsmallbg}{\strut\ \textcolor{wintext}{light green}\ } TSTR better by ${\leq}3\%$;\quad
\colorbox{tiebg}{\strut\ \textcolor{tietext}{yellow}\ } Within $+3\%$ tolerance;\quad
\colorbox{lossbg}{\strut\ \textcolor{losstext}{\textit{red}}\ } Gap ${>}3\%$.
\textbf{Run key:}\quad
$A$, $B$ = real datasets; $C$ = synthetic from $A$; $D$ = synthetic from $B$.
$A\!\to\!B$, $B\!\to\!A$ = TRTR; $D\!\to\!B$, $C\!\to\!A$ = TSTR.

\end{minipage}
\endgroup
\end{table*}

\subsection{Datasets and Evaluation Protocol}
\label{subsec:datasets}
We collect a benchmark of twelve real-world (nine multivariate and three univariate) time series~\cite{aksugift} spanning five qualitatively distinct domains: energy consumption~(\textit{BDG-2 Bear, Panther, Bull, Hog})~\citep{emami2023buildingsbench}, cloud infrastructure (\textit{Azure VM 2017, Borg 2011})~\citep{woo2023cloudops}, urban traffic (\textit{PEMS-04, PEMS-08})~\citep{jiang2023libcity}, climate (\textit{Subseasonal Precipitation, Subseasonal})~\citep{mouatadid2023subseasonalclimateusa} and residential energy (\textit{Residential PV Power, Residential Load Power})~\citep{moirai}. Within each domain we define source-matched \emph{sibling pairs} — datasets that share the same physical process, sensor modality, and temporal cadence so that the reference-guided generation procedure can extract a meaningful periodic scaffold from the source and apply it to synthesise training data for the target.

All experiments use a uniform input context of 96 time steps and a prediction horizon of 24 steps, evaluated by mean squared error (MSE) and mean absolute error (MAE). We consider three training protocols: \textbf{TRTR} (\textbf{T}rain on \textbf{R}eal, \textbf{T}est on \textbf{R}eal) serves as the gold-standard upper bound for each setting; \textbf{TSTR} (\textbf{T}rain on \textbf{S}ynthetic, \textbf{T}est on \textbf{R}eal) isolates the standalone value of synthetic data; and \textbf{TRSTR} (\textbf{T}rain on the union of \textbf{R}eal and \textbf{S}ynthetic, \textbf{T}est on \textbf{R}eal) measures augmentation benefit. To ensure that differences in TSTR versus TRTR cannot be attributed to corpus size, the synthetic corpus in every transfer setting is size-matched to the corresponding real corpus. Full dataset statistics, system specifications, sibling-pair definitions, and the characteristic period $P$ used for template extraction are summarised in Appendix \ref{app:dataset_details}.

We evaluate three forecasting backbones with deliberately distinct inductive biases: \textbf{iTransformer}~\citep{liu2024itransformer} ($\approx$1M parameters, self-attention over variates), \textbf{DLinear}~\citep{zeng2023dlinear} (4,673 parameters, explicit seasonal-residual decomposition), \textbf{S-Mamba}~\citep{wang2025smamba} ($\approx$1.1M parameters, state-space sequence modelling), \textbf{Moirai-small}~\cite{moirai}($\approx$13.8M parameters, a masked encoder-based time series foundational model). Evaluating across architectures with such different inductive biases provides a stringent test: if synthetic data genuinely captures domain structure rather than model-specific artefacts, its benefits should appear consistently across all model families.
 
\subsection{Q1: Reference-Guided Transfer as a Drop-in Replacement for Real Sibling Data}
\label{subsec:transfer}
We report TRTR and TSTR performance across all sibling pairs and backbones in Table~\ref{tab:regen_results_3pct}. \textbf{67\% of TSTR cells} (48 of 72) fall within the $\pm$3\% relative MSE band of their TRTR baseline — meaning that in two-thirds of settings, replacing real sibling data with \textsc{ReGEN}-generated series costs under 3\% in forecast accuracy, a margin within typical evaluation noise.
 
\textbf{Domain-wise variation.} Traffic is the strongest domain: on PEMS-04$\rightarrow$PEMS-08, \textsc{ReGEN} \emph{improves} over the real sibling for both iTransformer and DLinear ($-6.3$\% relative MSE); the reverse direction yields similar gains ($-9.4$\%, $-10.3$\%). The phase-aligned template is well-suited to traffic's regular intra-day periodicity, and temperature-broadened GP sampling extends coverage into rare congestion regimes. Cloud infrastructure (Azure VM / Borg) gains are smaller: aperiodic load spikes dominate several channels, leaving more variance for the GP residual under limited reference observations (see Appendix \ref{app:ve_by_covariate}). Residential PV is the weakest domain (DLinear gap: $+16.4$\%), as solar irradiance is locally variable and poorly approximated by a single periodic template.

\subsection{Q2: Synthetic Augmentation Improves Over Real-Only Training}
\label{subsec:augmentation}
We compares TRTR against TRSTR, training on the union of the real corpus and a size-matched synthetic corpus, across the same twelve datasets and three backbones in Table~\ref{tab:trtr_trstr_all_datasets}.

\begin{table}[!t]
\centering
\caption{TRTR vs.\ TRSTR across twelve datasets and three 
backbones. TRSTR adds a size-matched \textsc{ReGEN} synthetic 
corpus to the real training pool. Augmentation improves or 
matches TRTR in the majority of settings. Green indicates 
improvement; red indicates degradation. Lower is better 
($\downarrow$).}
\label{tab:trtr_trstr_all_datasets}
\fontsize{3.9}{4.4}\selectfont
\setlength{\tabcolsep}{0.8pt}
\renewcommand{\arraystretch}{0.82}
\resizebox{0.8\linewidth}{!}{%
\begin{tabular}{l c c c c c c c c c c c c}
\toprule
\multirow{3}{*}{\textbf{Dataset}} & \multicolumn{4}{c}{\textbf{iTransformer}} & \multicolumn{4}{c}{\textbf{DLinear}} & \multicolumn{4}{c}{\textbf{S-Mamba}} \\
\cmidrule(lr){2-5}\cmidrule(lr){6-9}\cmidrule(lr){10-13}
 & \multicolumn{2}{c}{\textbf{TRTR}} & \multicolumn{2}{c}{\textbf{TRSTR}} & \multicolumn{2}{c}{\textbf{TRTR}} & \multicolumn{2}{c}{\textbf{TRSTR}} & \multicolumn{2}{c}{\textbf{TRTR}} & \multicolumn{2}{c}{\textbf{TRSTR}} \\
\cmidrule(lr){2-13}
 & \textbf{MSE} & \textbf{MAE} & \textbf{MSE} & \textbf{MAE} & \textbf{MSE} & \textbf{MAE} & \textbf{MSE} & \textbf{MAE} & \textbf{MSE} & \textbf{MAE} & \textbf{MSE} & \textbf{MAE} \\
\midrule
Bear & 0.41 & 0.40 & \textcolor{textgreen}{0.38} & 0.40 & 0.41 & 0.39 & \textcolor{red}{0.42} & \textcolor{red}{0.43} & 0.42 & 0.34 & 0.42 & 0.34 \\
Panther & 0.32 & 0.36 & \textcolor{textgreen}{0.27} & \textcolor{textgreen}{0.35} & 0.29 & 0.36 & 0.29 & \textcolor{textgreen}{0.34} & 0.29 & 0.34 & 0.29 & \textcolor{textgreen}{0.33} \\
Azure VM & 0.88 & 0.41 & \textcolor{textgreen}{0.78} & \textcolor{textgreen}{0.30} & 0.90 & 0.44 & \textcolor{textgreen}{0.88} & \textcolor{textgreen}{0.42} & 0.90 & 0.41 & \textcolor{textgreen}{0.87} & \textcolor{textgreen}{0.40} \\
Borg & 0.58 & 0.49 & \textcolor{textgreen}{0.55} & \textcolor{textgreen}{0.40} & 0.59 & 0.54 & \textcolor{textgreen}{0.55} & 0.54 & 0.58 & 0.53 & \textcolor{textgreen}{0.54} & \textcolor{textgreen}{0.50} \\
PEMS-04 & 0.31 & 0.31 & \textcolor{textgreen}{0.25} & \textcolor{textgreen}{0.27} & 0.32 & 0.31 & \textcolor{textgreen}{0.30} & 0.31 & 0.30 & 0.28 & \textcolor{textgreen}{0.28} & \textcolor{red}{0.29} \\
PEMS-08 & 0.32 & 0.34 & \textcolor{textgreen}{0.27} & \textcolor{textgreen}{0.29} & 0.30 & 0.29 & 0.30 & \textcolor{textgreen}{0.28} & 0.29 & 0.30 & \textcolor{textgreen}{0.26} & \textcolor{textgreen}{0.29} \\
Bull & 0.33 & 0.40 & \textcolor{textgreen}{0.30} & \textcolor{textgreen}{0.35} & 0.36 & 0.39 & \textcolor{textgreen}{0.32} & \textcolor{textgreen}{0.35} & 0.33 & 0.40 & \textcolor{textgreen}{0.32} & 0.40 \\
Hog & 0.50 & 0.49 & \textcolor{textgreen}{0.47} & \textcolor{textgreen}{0.42} & 0.48 & 0.45 & \textcolor{textgreen}{0.47} & \textcolor{textgreen}{0.44} & 0.49 & 0.46 & \textcolor{textgreen}{0.43} & \textcolor{textgreen}{0.45} \\
Subseq. & 0.42 & 0.46 & \textcolor{red}{0.43} & \textcolor{red}{0.47} & 0.40 & 0.47 & \textcolor{red}{0.44} & \textcolor{red}{0.48} & 1.34 & 1.02 & \textcolor{textgreen}{1.08} & \textcolor{textgreen}{0.73} \\
Subseq. Prec. & 1.09 & 0.77 & \textcolor{textgreen}{0.92} & \textcolor{textgreen}{0.72} & 0.83 & 0.60 & \textcolor{textgreen}{0.79} & \textcolor{red}{0.62} & 0.76 & 0.57 & \textcolor{red}{0.78} & 0.57 \\
Res. PV & 0.25 & 0.22 & \textcolor{textgreen}{0.24} & \textcolor{red}{0.23} & 0.25 & 0.24 & \textcolor{red}{0.31} & \textcolor{red}{0.28} & 0.23 & 0.20 & \textcolor{red}{0.29} & \textcolor{red}{0.26} \\
Res. Load & 0.60 & 0.42 & \textcolor{textgreen}{0.54} & \textcolor{textgreen}{0.33} & 0.55 & 0.40 & \textcolor{red}{0.63} & \textcolor{red}{0.43} & 0.54 & 0.37 & 0.54 & 0.37 \\
\bottomrule
\end{tabular}%
}
\end{table}

\textbf{Augmentation helps broadly, but the effect is architecture-dependent.} For iTransformer and S-Mamba, TRSTR consistently matches or beats TRTR across the majority of datasets: both architectures benefit from the increased effective diversity of the combined corpus, absorbing the additional synthetic variation rather than overfitting the original distribution. DLinear shows the weakest and most mixed gains. This is mechanistically interpretable: DLinear's explicit seasonal-residual decomposition imposes a strong inductive bias that already extracts much of the periodic structure \textsc{ReGEN} synthesises. The model therefore has less capacity to benefit from additional variation in the periodic component, and the residual-level diversity added by the GP component can slightly confuse its residual-fitting head on some datasets.

The most practically significant augmentation gains appear where datasets are small and domain-specific. Subseasonal S-Mamba improves from an MSE of 1.34 (TRTR) to 1.02 (TRSTR) (23.9\% reduction) while iTransformer and DLinear remain stable. For Residential Load Power, augmentation yields consistent gains across all three backbones, suggesting that the cross-variate SCM coupling in the synthesised data supplies cross-channel patterns not fully captured by the limited real training corpus.


\subsection{Q3: ReGEN Produces Superior Training Signal In Comparison to Available Alternatives}
\label{subsec:full_corpus}
We benchmark \textsc{ReGEN} against two qualitatively distinct synthetic data generators: (1) \textbf{TimeGAN}~\cite{yoon2019timegan} — a reference-guided adversarial generator that trains a GAN directly on observed data, and (2) \textbf{CauKer}~\cite{xie2025cauker} — a reference-free, prior-based generator whose causal graph and kernel bank are sampled from domain-agnostic priors without conditioning on any real dataset. Table~\ref{tab:full_corpus_foundation_models} reports full-corpus TSTR results, where Moirai-small is pre-trained on each synthetic corpus and evaluated zero-shot on the pooled, left-out real benchmark.

\textbf{\textsc{ReGEN} Vs TimeGAN.} Both methods have access to 
the same observed data and identical corpus size budgets, so any 
performance gap reflects the \emph{quality} of what each method 
extracts from that reference. On Moirai, \textsc{ReGEN} reduces 
MSE by \textbf{41.1\%}. This advantage is broad: 
Table~\ref{tab:timegan_vs_regen_tstr} shows that \textsc{ReGEN} 
achieves lower MSE and MAE on \textbf{10 of 12 datasets}. The 
two exceptions (Bull, Hog) are datasets where the template stage 
explains less of the original signal variance 
(Appendix Table~\ref{tab:ve_by_covariate}), so TimeGAN's 
template-free design is less exposed to weak periodic fit.


\textbf{\textsc{ReGEN} Vs CauKer Vs TimeGAN.} CauKer's 
domain-agnostic design means that a corpus generated for cloud 
data is statistically indistinguishable from one generated for 
traffic data, making per-dataset TSTR comparisons methodologically 
unsound. We therefore pool all synthetic datasets from all three 
methods into a single full corpus and compare them in this regime, 
where CauKer's design assumption is respected. Moirai-small is 
pre-trained from scratch on each full corpus and evaluated 
zero-shot on the pooled real benchmark. \textsc{ReGEN} reduces 
Moirai-small MSE by \textbf{2.3\%}, demonstrating that 
conditioning generation on real reference data leaves a durable 
structural imprint that benefits downstream generalization beyond 
what domain-agnostic priors alone can provide.

\begin{table}[!t]
\centering
\caption{Dataset-wise iTransformer comparison. The TRTR block is shown as a neutral real-data reference. In the TSTR columns, \textcolor{textgreen}{green} marks the lower-error synthetic method between TimeGAN and \textsc{ReGeN} for each metric; lower is better.}
\label{tab:timegan_vs_regen_tstr}
\setlength{\tabcolsep}{3.0pt}
\renewcommand{\arraystretch}{0.92}
\begin{tabular}{l c c c c c c}
\toprule
\multirow{2}{*}{\textbf{Dataset}} & \multicolumn{2}{c}{\textbf{TRTR}} & \multicolumn{2}{c}{\textbf{TimeGAN}} & \multicolumn{2}{c}{\textbf{\textsc{ReGeN} (Ours)}} \\
\cmidrule(lr){2-3}\cmidrule(lr){4-5}\cmidrule(lr){6-7}
 & \textbf{MSE} & \textbf{MAE} & \textbf{MSE} & \textbf{MAE} & \textbf{MSE} & \textbf{MAE} \\
\midrule
Bear & 0.41 & 0.40 & \rowsecondbest{0.63} & \rowsecondbest{0.50} & \rowbest{0.41} & \rowbest{0.43} \\
Panther & 0.32 & 0.36 & \rowsecondbest{0.40} & \rowsecondbest{0.45} & \rowbest{0.36} & \rowbest{0.38} \\
Azure VM & 0.88 & 0.41 & \rowsecondbest{1.04} & \rowsecondbest{0.51} & \rowbest{0.90} & \rowbest{0.45} \\
Borg & 0.58 & 0.49 & \rowsecondbest{0.85} & \rowsecondbest{0.69} & \rowbest{0.59} & \rowbest{0.50} \\
PEMS-04 & 0.31 & 0.31 & \rowsecondbest{0.45} & \rowsecondbest{0.45} & \rowbest{0.37} & \rowbest{0.38} \\
PEMS-08 & 0.32 & 0.34 & \rowsecondbest{0.40} & \rowsecondbest{0.38} & \rowbest{0.30} & \rowbest{0.29} \\
Bull & 0.33 & 0.40 & \rowbest{0.32} & \rowbest{0.36} & \rowsecondbest{0.37} & \rowsecondbest{0.40} \\
Hog & 0.50 & 0.49 & \rowbest{0.47} & \rowsecondbest{0.49} & \rowsecondbest{0.50} & \rowbest{0.48} \\
Subseasonal & 0.42 & 0.46 & \rowsecondbest{0.42} & \rowsecondbest{0.45} & \rowbest{0.40} & \rowbest{0.43} \\
Sub. precip. & 1.09 & 0.77 & \rowsecondbest{1.11} & \rowsecondbest{0.80} & \rowbest{1.01} & \rowbest{0.73} \\
Residential PV & 0.25 & 0.22 & \rowsecondbest{0.27} & \rowbest{0.20} & \rowbest{0.26} & \rowbest{0.20} \\
Residential Load & 0.60 & 0.42 & \rowsecondbest{0.71} & \rowsecondbest{0.49} & \rowbest{0.54} & \rowbest{0.38} \\
\bottomrule
\end{tabular}%
\end{table}


\begin{table}[t!]
\centering
\caption{Full-corpus TSTR evaluation with Moirai-small, comparing 
the different synthetic corpora used for pre-training. Lower MSE, 
MAE, MASE, and WQL are better. Best results in 
\textcolor{textgreen}{green}.}
\label{tab:full_corpus_foundation_models}
\setlength{\tabcolsep}{6pt}
\renewcommand{\arraystretch}{1.12}
\begin{tabular}{l c c c c}
\toprule
\textbf{Synthetic Corpus} & \textbf{MSE} & \textbf{MAE} & \textbf{MASE} & \textbf{WQL} \\
\midrule
TimeGAN & 392.46 & 8.38 & 1.45 & 0.23 \\
CauKer & 236.7 & 6.64 & 1.11 & 0.17 \\
\textsc{ReGeN} (Ours) & \textcolor{textgreen}{231.14} & \textcolor{textgreen}{5.86} & \textcolor{textgreen}{0.98} & \textcolor{textgreen}{0.17} \\
\bottomrule
\end{tabular}
\end{table}

\subsection{Appendix C: Ablation and Diagnostic Summary}
\label{subsec:ablation}
Appendix~C consolidates five complementary results that clarify \emph{why} \textsc{ReGeN} works. \ding{182} Table~\ref{tab:ve_by_covariate} shows that the phase-aligned template already captures a large share of variance in the strongest domains, especially traffic and residential energy, while also revealing the lower-VE settings in which the residual model must carry more of the burden. \ding{183} Figure~\ref{fig:tsne_grid_appendix} shows that the synthetic samples recover the geometry of the real data closely enough to preserve domain-level clustering structure. \ding{184} Figure~\ref{fig:psd_grid_appendix} substantiates that the synthetic series largely match the real datasets' dominant spectral peaks and low-frequency decay, supporting strong frequency-domain fidelity. \ding{185} Table~\ref{tab:tstr_no_residual_ablation} shows that removing the deep-kernel GP residual increases mean iTransformer TSTR MSE by 11.0\% across the reported datasets, confirming that residual uncertainty modelling contributes materially to downstream performance. \ding{186} Table~\ref{tab:tstr_no_scm_ablation} shows that removing SCM-based mixing increases mean iTransformer TSTR MSE by 6.0\%, indicating that directed cross-variate coupling provides an additional, consistent gain. Taken together, these results show that \textsc{ReGeN}'s gains are not driven by one component alone: they emerge from a strong periodic scaffold, realistic geometric and spectral fidelity, and measurable benefits from both residual modelling and SCM-based mixing.
\section{Conclusion}
We presented \textsc{ReGeN}, a reference-guided generative pipeline that decomposes observed multivariate sequences into a phase-aligned periodic template, a deep-kernel GP residual, and a fitted structural causal model. By grounding all three components in real domain observations, it produces synthetic data that inherits the periodic morphology, local uncertainty, and cross-variable coupling of the target domain without requiring large training corpora. Across twelve datasets and five domains, \textsc{ReGeN}-synthesized data substitutes for real sibling data within a 3\% MSE margin in two-thirds of transfer settings, outperforms real-data transfer in strongly periodic domains, and yields consistent augmentation gains for attention- and state-space-based architectures. 

\textbf{Limitations.} Our current evaluation has three main limitations. First, SCM-based mixing is not uniformly beneficial: in low-data, higher-dimensional settings such as Hog, graph estimation can become noisy and the induced cross-variate coupling may hurt performance. Second, due to computational constraints, we report full-corpus pretraining results for Moirai-small only; extending this comparison to additional foundation models would require training each architecture from scratch on each synthetic corpus. Third, our component ablations are reported only for iTransformer, so while they support the design rationale, they do not yet establish that the same component-level effects hold equally strongly across all downstream model families.
{
\small
\bibliographystyle{plainnat}
\bibliography{main_bib}

@misc{moirai,
      title={Unified Training of Universal Time Series Forecasting Transformers}, 
      author={Gerald Woo and Chenghao Liu and Akshat Kumar and Caiming Xiong and Silvio Savarese and Doyen Sahoo},
      year={2024},
      eprint={2402.02592},
      archivePrefix={arXiv},
      primaryClass={cs.LG},
      url={https://arxiv.org/abs/2402.02592}, 
}

@misc{chronos,
      title={Chronos: Learning the Language of Time Series}, 
      author={Abdul Fatir Ansari and Lorenzo Stella and Caner Turkmen and Xiyuan Zhang and Pedro Mercado and Huibin Shen and Oleksandr Shchur and Syama Sundar Rangapuram and Sebastian Pineda Arango and Shubham Kapoor and Jasper Zschiegner and Danielle C. Maddix and Hao Wang and Michael W. Mahoney and Kari Torkkola and Andrew Gordon Wilson and Michael Bohlke-Schneider and Yuyang Wang},
      year={2024},
      eprint={2403.07815},
      archivePrefix={arXiv},
      primaryClass={cs.LG},
      url={https://arxiv.org/abs/2403.07815}, 
}

@inproceedings{taga2025timepfn,
  title={Timepfn: Effective multivariate time series forecasting with synthetic data},
  author={Taga, Ege Onur and Ildiz, Muhammed Emrullah and Oymak, Samet},
  booktitle={Proceedings of the AAAI conference on artificial intelligence},
  volume={39},
  pages={20761--20769},
  year={2025}
}

@article{cleveland1990stl,
  title={STL: A Seasonal-Trend Decomposition Procedure Based on Loess},
  author={Cleveland, Robert B. and Cleveland, William S. and McRae, Jean E. and Terpenning, Irma},
  journal={Journal of Official Statistics},
  volume={6},
  number={1},
  pages={3--73},
  year={1990}
}

@article{roberts2013gaussian,
  title={Gaussian Processes for Time-Series Modelling},
  author={Roberts, Stephen and Osborne, Michael and Ebden, Mark and Reece, Steven and Gibson, Nicholas and Aigrain, Suzanne},
  journal={Philosophical Transactions of the Royal Society A: Mathematical, Physical and Engineering Sciences},
  volume={371},
  number={1984},
  pages={20110550},
  year={2013},
  doi={10.1098/rsta.2011.0550}
}

@inproceedings{wilson2016deep,
  title={Deep Kernel Learning},
  author={Wilson, Andrew Gordon and Hu, Zhiting and Salakhutdinov, Ruslan and Xing, Eric P.},
  booktitle={Proceedings of the 19th International Conference on Artificial Intelligence and Statistics},
  series={Proceedings of Machine Learning Research},
  volume={51},
  pages={370--378},
  year={2016},
  url={https://proceedings.mlr.press/v51/wilson16.html}
}

@article{alshedivat2017learning,
  title={Learning Scalable Deep Kernels with Recurrent Structure},
  author={Al-Shedivat, Maruan and Wilson, Andrew Gordon and Saatchi, Yunus and Hu, Zhiting and Xing, Eric P.},
  journal={Journal of Machine Learning Research},
  volume={18},
  number={82},
  pages={1--37},
  year={2017},
  url={https://jmlr.org/papers/v18/16-498.html}
}

@book{rasmussen2006gaussian,
  title={Gaussian Processes for Machine Learning},
  author={Rasmussen, Carl Edward and Williams, Christopher K. I.},
  year={2006},
  publisher={MIT Press}
}

@article{hochreiter1997long,
  title={Long Short-Term Memory},
  author={Hochreiter, Sepp and Schmidhuber, J{\"u}rgen},
  journal={Neural Computation},
  volume={9},
  number={8},
  pages={1735--1780},
  year={1997},
  doi={10.1162/neco.1997.9.8.1735}
}

@book{peters2017elements,
  title={Elements of Causal Inference: Foundations and Learning Algorithms},
  author={Peters, Jonas and Janzing, Dominik and Sch{\"o}lkopf, Bernhard},
  year={2017},
  publisher={MIT Press}
}

@misc{xie2025cauker,
  title={{CauKer}: Classification Time Series Foundation Models Can Be Pretrained on Synthetic Data Only},
  author={Xie, Shifeng and Feofanov, Vasilii and Alonso, Marius and Odonnat, Ambroise and Zhang, Jianfeng and Palpanas, Themis and Zan, Lei and Pan, Lujia and Zhang, Keli and Redko, Ievgen},
  year={2025},
  eprint={2508.02879},
  archivePrefix={arXiv},
  url={https://arxiv.org/abs/2508.02879}
}

@inproceedings{yoon2019timegan,
  title={Time-series Generative Adversarial Networks},
  author={Yoon, Jinsung and Jarrett, Daniel and van der Schaar, Mihaela},
  booktitle={Advances in Neural Information Processing Systems 32},
  year={2019},
  url={https://papers.nips.cc/paper/8789-time-series-generative-adversarial-networks}
}

@inproceedings{dooley2023forecastpfn,
  title={ForecastPFN: Synthetically-Trained Zero-Shot Forecasting},
  author={Dooley, Samuel and Khurana, Gurnoor Singh and Mohapatra, Chirag and Naidu, Siddartha Venkat and White, Colin},
  booktitle={Advances in Neural Information Processing Systems 36},
  year={2023},
  url={https://proceedings.neurips.cc/paper_files/paper/2023/hash/0731f0e65559059eb9cd9d6f44ce2dd8-Abstract-Conference.html}
}

@inproceedings{darlow2023tsmix,
  title={TSMix: Time Series Data Augmentation by Mixing Sources},
  author={Darlow, Luke and Asenov, Martin and Joosen, Artjom and Deng, Qiwen and Wang, Jianfeng and Barker, Adam David},
  booktitle={Proceedings of the 3rd Workshop on Machine Learning and Systems},
  year={2023},
  pages={109--114},
  doi={10.1145/3578356.3592584}
}

@misc{aggarwal2023mixuppp,
  title={Embarrassingly Simple MixUp for Time-series},
  author={Aggarwal, Karan and Srivastava, Jaideep},
  year={2023},
  eprint={2304.04271},
  archivePrefix={arXiv},
  primaryClass={cs.LG},
  url={https://arxiv.org/abs/2304.04271}
}

@inproceedings{aksugift,
  title={GIFT-Eval: A Benchmark for General Time Series Forecasting Model Evaluation},
  author={Aksu, Taha and Woo, Gerald and Liu, Juncheng and Liu, Xu and Liu, Chenghao and Savarese, Silvio and Xiong, Caiming and Sahoo, Doyen},
  booktitle={NeurIPS Workshop on Time Series in the Age of Large Models},
  year={2024}
}

@article{mogren2016c,
  title={C-RNN-GAN: Continuous recurrent neural networks with adversarial training},
  author={Mogren, Olof},
  journal={arXiv preprint arXiv:1611.09904},
  year={2016}
}

@article{esteban2017real,
  title={Real-valued (medical) time series generation with recurrent conditional gans},
  author={Esteban, Crist{\'o}bal and Hyland, Stephanie L and R{\"a}tsch, Gunnar},
  journal={arXiv preprint arXiv:1706.02633},
  year={2017}
}

@misc{oreshkin2026sarsim0,
  title={Zero-shot Forecasting by Simulation Alone},
  author={Oreshkin, Boris N. and Jauhari, Mayank and Selvam, Ravi Kiran and Wolff, Malcolm and Pan, Wenhao and Ramasubramanian, Shankar and Olivares, Kin G. and Konstantinova, Tatiana and Potapczynski, Andres and Cao, Mengfei and Efimov, Dmitry and Mahoney, Michael W. and Wilson, Andrew G.},
  year={2026},
  eprint={2601.00970},
  archivePrefix={arXiv},
  primaryClass={cs.LG},
  url={https://arxiv.org/abs/2601.00970}
}

@inproceedings{emami2023buildingsbench,
  title={BuildingsBench: A Large-Scale Dataset of 900K Buildings and Benchmark for Short-Term Load Forecasting},
  author={Emami, Patrick and Sahu, Abhijeet and Graf, Peter},
  booktitle={NeurIPS 2023 Track on Datasets and Benchmarks},
  year={2023},
  url={https://openreview.net/forum?id=c5rqd6PZn6}
}

@inproceedings{mouatadid2023subseasonalclimateusa,
  title={SubseasonalClimateUSA: A Dataset for Subseasonal Forecasting and Benchmarking},
  author={Mouatadid, Soukayna and Orenstein, Paulo and Flaspohler, Genevieve Elaine and Oprescu, Miruna and Cohen, Judah and Wang, Franklyn and Knight, Sean Edward and Geogdzhayeva, Maria and Levang, Samuel James and Fraenkel, Ernest and Mackey, Lester},
  booktitle={NeurIPS 2023 Track on Datasets and Benchmarks},
  year={2023},
  url={https://openreview.net/forum?id=pWkrU6raMt}
}

@article{jiang2023libcity,
  title={LibCity: A Unified Library Towards Efficient and Comprehensive Urban Spatial-Temporal Prediction},
  author={Jiang, Jiawei and Han, Chengkai and Jiang, Wenjun and Zhao, Wayne Xin and Wang, Jingyuan},
  journal={arXiv preprint arXiv:2304.14343},
  year={2023},
  url={https://arxiv.org/abs/2304.14343}
}

@article{woo2023cloudops,
  title={Pushing the Limits of Pre-training for Time Series Forecasting in the CloudOps Domain},
  author={Woo, Gerald and Liu, Chenghao and Kumar, Akshat and Sahoo, Doyen},
  journal={arXiv preprint arXiv:2310.05063},
  year={2023},
  url={https://arxiv.org/abs/2310.05063}
}

@inproceedings{liu2024itransformer,
  title={iTransformer: Inverted Transformers Are Effective for Time Series Forecasting},
  author={Liu, Yong and Hu, Tengge and Zhang, Haoran and Wu, Haixu and Wang, Shiyu and Ma, Lintao and Long, Mingsheng},
  booktitle={The Twelfth International Conference on Learning Representations},
  year={2024},
  url={https://openreview.net/forum?id=JePfAI8fah}
}

@article{wang2025smamba,
  title={Is Mamba Effective for Time Series Forecasting?},
  author={Wang, Zihan and Kong, Fanheng and Feng, Shi and Wang, Ming and Yang, Xiaocui and Zhao, Han and Wang, Daling and Zhang, Yifei},
  journal={Neurocomputing},
  volume={619},
  pages={129178},
  year={2025},
  doi={10.1016/j.neucom.2024.129178},
  url={https://www.sciencedirect.com/science/article/pii/S0925231224019490}
}

@inproceedings{zeng2023dlinear,
  title={Are Transformers Effective for Time Series Forecasting?},
  author={Zeng, Ailing and Chen, Muxi and Zhang, Lei and Xu, Qiang},
  booktitle={Proceedings of the AAAI Conference on Artificial Intelligence},
  volume={37},
  pages={11121--11128},
  year={2023},
  doi={10.1609/aaai.v37i9.26317}
}

@article{runge2019detecting,
  title={Detecting and quantifying causal associations in large nonlinear time series datasets},
  author={Runge, Jakob and Nowack, Peer and Kretschmer, Marlene and Flaxman, Seth and Sejdinovic, Dino},
  journal={Science Advances},
  volume={5},
  number={11},
  pages={eaau4996},
  year={2019},
  doi={10.1126/sciadv.aau4996}
}

@inproceedings{pamfil2020dynotears,
  title={DYNOTEARS: Structure Learning from Time-Series Data},
  author={Pamfil, Roxana and Sriwattanaworachai, Nisara and Desai, Shaan and Pilgerstorfer, Philip and Beaumont, Paul and Georgatzis, Konstantinos and Aragam, Bryon},
  booktitle={Proceedings of the Twenty Third International Conference on Artificial Intelligence and Statistics},
  pages={1595--1605},
  year={2020}
}

@article{hyvarinen2010estimation,
  title={Estimation of a Structural Vector Autoregression Model Using Non-Gaussianity},
  author={Hyv{\"a}rinen, Aapo and Zhang, Kun and Shimizu, Shohei and Hoyer, Patrik O.},
  journal={Journal of Machine Learning Research},
  volume={11},
  pages={1709--1731},
  year={2010}
}

@article{granger1969investigating,
  title={Investigating Causal Relations by Econometric Models and Cross-spectral Methods},
  author={Granger, C. W. J.},
  journal={Econometrica},
  volume={37},
  number={3},
  pages={424--438},
  year={1969},
  doi={10.2307/1912791}
}

@article{breiman2001random,
  title={Random Forests},
  author={Breiman, Leo},
  journal={Machine Learning},
  volume={45},
  pages={5--32},
  year={2001},
  doi={10.1023/A:1010933404324}
}

@article{maaten2008visualizing,
  title={Visualizing Data using t-SNE},
  author={van der Maaten, Laurens and Hinton, Geoffrey},
  journal={Journal of Machine Learning Research},
  volume={9},
  pages={2579--2605},
  year={2008}
}

@article{welch1967use,
  title={The Use of Fast Fourier Transform for the Estimation of Power Spectra: A Method Based on Time Averaging Over Short, Modified Periodograms},
  author={Welch, Peter D.},
  journal={IEEE Transactions on Audio and Electroacoustics},
  volume={15},
  number={2},
  pages={70--73},
  year={1967},
  doi={10.1109/TAU.1967.1161901}
}

@article{spearman1904proof,
  title={The Proof and Measurement of Association Between Two Things},
  author={Spearman, Charles},
  journal={The American Journal of Psychology},
  volume={15},
  number={1},
  pages={72--101},
  year={1904},
  doi={10.2307/1412159}
}
}

\newpage
\begin{appendices}

\newcommand{\pspdf}[2]{\includegraphics[page=#2,width=0.155\linewidth]{#1}}
\newcommand{\pspdfthree}[2]{\includegraphics[page=#2,width=0.30\textwidth]{#1}}
\newcommand{\pspdffour}[2]{\includegraphics[page=#2,width=0.225\textwidth]{#1}}

\clearpage
\section{Dataset and System Details}
\label{app:dataset_details}

\subsection{Experimental System Details}
\label{app:system_details}
All experiments reported in the main paper and supplementary material were run on the same machine. The hardware consisted of an NVIDIA A10G GPU with 22.1~GB of VRAM, an AMD EPYC 7R32 CPU with 4 cores and 8 threads, 32~GB of system RAM, and a 242~GB SSD. The software environment used Ubuntu 24.04.4~LTS. These specifications cover both synthetic-data generation and downstream forecasting experiments.

\subsection{Dataset Details}

\noindent\begin{minipage}{\linewidth}
\centering
\captionof{table}{Twelve real-world datasets used in evaluation, spanning energy, cloud, traffic, and climate, with summary statistics for sampling frequency, characteristic template period \(P\), series count, targets, and covariates.}
\label{tab:dataset_summary}
\small
\setlength{\tabcolsep}{4pt}
\renewcommand{\arraystretch}{1.08}
\resizebox{0.94\linewidth}{!}{%
\begin{tabular}{lllc c c c c}
\toprule
\textbf{Dataset} & \textbf{Source} & \textbf{Domain} & \textbf{Frequency} & \textbf{Template \(P\)} & \textbf{\# Time Series} & \textbf{\# Targets} & \textbf{\# Covariates} \\
\midrule
BDG-2 Bear & BuildingsBench~\citep{emami2023buildingsbench} & Energy & H & 24 & 91 & 1 & 0 \\
BDG-2 Panther & BuildingsBench~\citep{emami2023buildingsbench} & Energy & H & 24 & 105 & 1 & 0 \\
Azure VM Traces 2017 & CloudOpsTSF~\citep{woo2023cloudops} & Cloud & 5T & 288 & 10,000 & 1 & 2 \\
Borg Cluster Data 2011 & CloudOpsTSF~\citep{woo2023cloudops} & Cloud & 5T & 288 & 10,000 & 2 & 5 \\
PEMS-04 & LibCity~\citep{jiang2023libcity} & Traffic & 5T & 288 & 307 & 3 & 0 \\
PEMS-08 & LibCity~\citep{jiang2023libcity} & Traffic & 5T & 288 & 170 & 3 & 0 \\
BDG-2 Bull & BuildingsBench~\citep{emami2023buildingsbench} & Energy & H & 24 & 41 & 1 & 3 \\
BDG-2 Hog & BuildingsBench~\citep{emami2023buildingsbench} & Energy & H & 24 & 24 & 1 & 5 \\
Subseasonal & SubseasonalClimateUSA~\citep{mouatadid2023subseasonalclimateusa} & Climate & D & 365 & 862 & 4 & 0 \\
Subseasonal Precipitation & SubseasonalClimateUSA~\citep{mouatadid2023subseasonalclimateusa} & Climate & D & 365 & 862 & 1 & 0 \\
Residential PV Power & LOTSA\_Others~\citep{moirai} & Energy & T & 1440 & 233 & 3 & 0 \\
Residential Load Power & LOTSA\_Others~\citep{moirai} & Energy & T & 1440 & 271 & 3 & 0 \\
\bottomrule
\end{tabular}%
}
\end{minipage}

\noindent\textbf{Characteristic Period.} \(P\) denotes the characteristic period length, in time steps, used for phase alignment in the template-extraction stage. We use one full daily cycle for the sub-daily datasets and one climatological yearly cycle for the daily climate datasets, so the frequency codes map as follows: \(\mathrm{H} \mapsto P=24\), \(\mathrm{5T} \mapsto P=288\), \(\mathrm{T} \mapsto P=1440\), and \(\mathrm{D} \mapsto P=365\).

The 12 datasets in Table~\ref{tab:dataset_summary} are grouped into six source-matched pairs so that each pair shares a common collection setting or application domain while still differing in scale, dimensionality, or predictand. We briefly summarize those six pairings here because they anchor the cross-dataset comparisons used throughout the main paper.

\noindent\textbf{BuildingsBench: Bear and Panther.} Bear and Panther are hourly building-energy datasets from BuildingsBench, both focused on single-target load forecasting without additional covariates. They form a clean univariate pair for evaluating whether a method can transfer across buildings that share broad consumption rhythms but differ in occupancy patterns, control policies, and building-specific demand variability.

\noindent\textbf{CloudOpsTSF: Azure VM Traces 2017 and Borg Cluster Data 2011.} Azure VM Traces 2017 and Borg Cluster Data 2011 represent cloud-resource monitoring at 5-minute resolution. This pair is useful because both datasets capture operational infrastructure telemetry, but Azure is a simpler single-target setting with two covariates, whereas Borg is multivariate and more heterogeneous, making the pair a direct test of how methods scale from lighter to richer cloud traces.

\noindent\textbf{LibCity: PEMS-04 and PEMS-08.} PEMS-04 and PEMS-08 are traffic datasets from LibCity with 5-minute sampling and three target channels per sensor. They provide a matched transportation pair in which both tasks exhibit strong daily and weekly rhythms, while differing in network size and sensor topology, so they probe whether a model preserves structured periodicity under varying spatial scales.

\noindent\textbf{BuildingsBench: Bull and Hog.} Bull and Hog return to BuildingsBench, but now in covariate-rich hourly settings rather than the univariate Bear/Panther case. Because both datasets model building demand with exogenous drivers, this pair helps separate performance gains from simple periodic load reconstruction versus the harder problem of handling auxiliary variables that may shift or weaken the dominant seasonal pattern.

\noindent\textbf{SubseasonalClimateUSA: Subseasonal and Subseasonal Precipitation.} Subseasonal and Subseasonal Precipitation come from the same climate benchmark and both aggregate daily measurements across 862 series, but they differ sharply in target dimensionality. The pair therefore isolates how a method behaves when the broader meteorological setting is held fixed while the predictive task changes from a four-target subseasonal forecasting problem to a precipitation-only version with a narrower signal profile.

\noindent\textbf{LOTSA Others: Residential PV Power and Residential Load Power.} Residential PV Power and Residential Load Power are minute-level energy datasets drawn from the LOTSA collection. They form a natural household-energy pair because both reflect residential behavior at fine temporal resolution, yet PV generation is dominated by solar forcing while load reflects human activity and appliance usage, giving a useful contrast between externally driven and behavior-driven dynamics.

\clearpage
\section{Consensus DAG Estimation}
\label{app:dag_estimation}

The graph $\mathcal{G}$ used in the SCM mixing stage is estimated directly from the real multivariate data through a consensus causal-discovery pipeline. We avoid relying on a single discovery routine because different estimators emphasize different dependence structures and can behave differently across datasets. Instead, we infer candidate edges with four complementary procedures: PCMCI~\citep{runge2019detecting}, DYNOTEARS~\citep{pamfil2020dynotears}, VARLiNGAM~\citep{hyvarinen2010estimation}, and a non-linear Granger-style test based on random-forest feature importance~\citep{granger1969investigating,breiman2001random}. The final graph retains only edges that recur across the dataset and receive support from multiple methods.

\noindent\textbf{Per-series preprocessing.} For each real series $s$, we stack all available variates into a single multivariate trajectory
\begin{equation}
\mathbf{z}^{(s)}_t = \bigl(z^{(s)}_{1,t},\dots,z^{(s)}_{C,t}\bigr)^\top, \qquad t=1,\dots,T_s,
\end{equation}
placing dynamic covariates before the target variates. Missing values are imputed by linear interpolation, with forward/backward filling at the boundaries when needed. Each variate is then standardized independently,
\begin{equation}
\widetilde z^{(s)}_{c,t} = \frac{z^{(s)}_{c,t} - \mu^{(s)}_c}{\sigma^{(s)}_c + \varepsilon},
\end{equation}
so that discovery is driven by temporal dependence rather than raw scale. Here $C$ is the total number of variates in the multivariate series and $T_s$ is its observed length.

\noindent\textbf{Candidate edge discovery.} We run all four discovery methods on each standardized series over a fixed lag window $\ell \in \{0,1,\dots,L_{\max}\}$. For a given series and method, a candidate edge is indexed by an ordered triple $(u,v,\ell)$, where $u \in \{1,\dots,C\}$ is the source variate, $v \in \{1,\dots,C\}$ is the destination variate, and $\ell$ is the lag relating source time $t-\ell$ to destination time $t$. Thus, $\ell=0$ denotes an instantaneous or contemporaneous relation, whereas $\ell>0$ denotes a delayed relation. Every method returns a set of candidate relations of the form
\begin{equation}
\widetilde z^{(s)}_{u,t-\ell} \rightarrow \widetilde z^{(s)}_{v,t}
\end{equation}
together with a raw edge-strength score $a^{(s,m)}_{u,v,\ell}$, where $m \in \{1,\dots,4\}$ indexes the discovery method. We keep both contemporaneous and positive-lag edges at this stage. However, self-edges with $u=v$ are removed before the final consensus graph is constructed. The reason is architectural rather than causal: the periodic template stage already carries the dominant self-lag or autoregressive structure, so re-inserting self-links into the SCM mixing graph would double-count that dependence, degrade signal quality, and typically reduce fidelity rather than improve it.

\noindent\textbf{Aggregation across series.} The raw scores produced by different algorithms are not directly comparable, so we normalize them within each method and series before aggregation. Let $\mathcal{E}^{(s,m)}$ denote the set of candidate edges returned by method $m$ on series $s$. We define the normalized score
\begin{equation}
\widehat a^{(s,m)}_{u,v,\ell} =
\begin{cases}
\dfrac{\left|a^{(s,m)}_{u,v,\ell}\right|}{\max\limits_{(i,j,r)\in\mathcal{E}^{(s,m)}} \left|a^{(s,m)}_{i,j,r}\right|}, & (u,v,\ell) \in \mathcal{E}^{(s,m)},\\[1.0em]
0, & \text{otherwise.}
\end{cases}
\end{equation}
where $(i,j,r)$ is simply a dummy edge index ranging over all candidate source variates $i$, destination variates $j$, and lags $r$ returned by method $m$ on series $s$. This rescales all detected edges for a given method/series pair into $[0,1]$ while preserving their relative ordering. We then aggregate each edge across the $S$ real series through two summaries. First, its frequency of occurrence under method $m$ is
\begin{equation}
f^{(m)}_{u,v,\ell} = \frac{1}{S} \sum_{s=1}^{S} \mathbf{1}\!\left[(u,v,\ell) \in \mathcal{E}^{(s,m)}\right].
\end{equation}
Second, its mean normalized strength over the series in which it appears is
\begin{equation}
\bar a^{(m)}_{u,v,\ell} =
\frac{\sum_{s=1}^{S} \widehat a^{(s,m)}_{u,v,\ell}}
{\max\!\left(1,\sum_{s=1}^{S} \mathbf{1}\!\left[(u,v,\ell) \in \mathcal{E}^{(s,m)}\right]\right)}.
\end{equation}
These two quantities separate stability across series from within-method edge magnitude, yielding a dataset-level summary for each method rather than a separate graph for every series. At this point the edge-lag triple $(u,v,\ell)$ is still represented by \emph{method-specific} summaries $\bar a^{(m)}_{u,v,\ell}$ rather than by a single pooled score. In other words, after this step there can still be up to four aggregated strengths for the same $(u,v,\ell)$, one for each discovery method.

\noindent\textbf{Consensus graph construction.} A vote is counted for an edge only after two levels of filtering. First, a particular discovery method $m$ casts a vote for $(u,v,\ell)$ only if that relation appears in more than a threshold fraction of the dataset. In our implementation, this per-method support condition is
\begin{equation}
f^{(m)}_{u,v,\ell} > \tau_{\mathrm{freq}}, \qquad \tau_{\mathrm{freq}} = 0.2.
\end{equation}
Equivalently, the edge must appear in more than 20\% of the series for that method. Defining the per-method vote indicator as
\begin{equation}
b^{(m)}_{u,v,\ell} = \mathbf{1}\!\left[f^{(m)}_{u,v,\ell} > \tau_{\mathrm{freq}}\right],
\end{equation}
the second requirement is cross-method agreement: an edge is kept only if at least two of the four discovery methods vote for it,
\begin{equation}
\sum_{m=1}^{4} b^{(m)}_{u,v,\ell} \ge 2.
\end{equation}
Thus, an edge enters the final consensus graph only if \emph{both} conditions hold: (a) for a given method it appears in more than $\tau_{\mathrm{freq}}$ of the dataset, and (b) at least two methods vote for that same edge. For edges that survive this voting stage, we then pool the method-specific strengths into a single final consensus score,
\begin{equation}
w_{u,v,\ell} = \frac{\sum_{m=1}^{4} b^{(m)}_{u,v,\ell} \, \bar a^{(m)}_{u,v,\ell}}{\sum_{m=1}^{4} b^{(m)}_{u,v,\ell}},
\qquad \text{defined whenever } \sum_{m=1}^{4} b^{(m)}_{u,v,\ell} \ge 2.
\end{equation}
This is the final per-edge, per-lag coefficient used downstream in SCM mixing. Subject to the additional constraint $u \neq v$ that removes self-links from downstream SCM mixing, we then collect all retained lags for each parent-child pair $(u,v)$ into the admissible lag set
\begin{equation}
\mathcal{L}_{u\to v} = \left\{\ell \in \{0,1,\dots,L_{\max}\} : \sum_{m=1}^{4} b^{(m)}_{u,v,\ell} \ge 2 \right\}.
\end{equation}
The resulting consensus graph therefore allows both instantaneous ($\ell=0$) and delayed ($\ell>0$) cross-variate relations, while leaving self-temporal structure to the periodic template and residual components described in Section~\ref{sec:method}.

\clearpage
\section{Ablation Studies}
\label{app:ablation}

To keep the main paper compact, we collect the full variance, geometric, spectral, residual, and SCM-mixing ablations below.

\subsection{Variance Explained by Covariate}
\label{app:ve_by_covariate}

\begin{center}
\captionof{table}{Variance explained (VE, \%) for each dataset and each available channel, together with the per-dataset average across available channels. Higher values indicate that the phase-aligned template explains a larger fraction of the observed signal variance.}
\label{tab:ve_by_covariate}
\small
\setlength{\tabcolsep}{4pt}
\renewcommand{\arraystretch}{1.08}
\resizebox{0.94\linewidth}{!}{%
\begin{tabular}{lccccccc}
\toprule
\textbf{Dataset} & \textbf{Covariate 1} & \textbf{Covariate 2} & \textbf{Covariate 3} & \textbf{Covariate 4} & \textbf{Covariate 5} & \textbf{Covariate 6} & \textbf{Average} \\
\midrule
Bear & 57.3 & -- & -- & -- & -- & -- & 57.3 \\
Panther & 39.7 & -- & -- & -- & -- & -- & 39.7 \\
Azure VM & 9.1 & 25.0 & 14.5 & -- & -- & -- & 16.2 \\
Borg & 59.3 & 15.2 & 14.8 & 0.7 & 35.4 & 59.4 & 30.8 \\
PEMS-04 & 91.9 & 84.7 & 65.4 & -- & -- & -- & 80.7 \\
PEMS-08 & 91.2 & 84.7 & 65.1 & -- & -- & -- & 80.3 \\
Bull & 14.8 & 0.9 & 3.4 & 2.1 & -- & -- & 5.3 \\
Hog & 3.2 & 0.9 & 0.3 & 1.6 & 8.8 & 1.7 & 2.8 \\
Subseasonal & 30.3 & 97.1 & 97.0 & 97.1 & -- & -- & 80.4 \\
Subseasonal Prec. & 32.6 & -- & -- & -- & -- & -- & 32.6 \\
Res. PV & 88.7 & 88.3 & 88.8 & -- & -- & -- & 88.6 \\
Res. Load & 88.1 & 92.0 & 93.6 & -- & -- & -- & 91.2 \\
\bottomrule
\end{tabular}%
}
\end{center}

Table~\ref{tab:ve_by_covariate} reports the variance explained (VE) by the phase-aligned periodic template for every dataset and channel. The values span a wide range, from below 1\% on individual Hog and Bull channels to above 90\% on PEMS and the residential benchmarks, reflecting genuine heterogeneity in how strongly periodic structure dominates across domains. Three broad tiers are visible. High-VE datasets, namely PEMS-04, PEMS-08, Residential PV Power, Residential Load Power, and the non-precipitation Subseasonal channels, have averages above 80\%, meaning the template captures most of the signal variance before residual modelling. Moderate-VE datasets, including Bear, Panther, Subseasonal Precipitation, and Borg, sit between roughly 30\% and 60\% on average, so the template provides a meaningful but incomplete scaffold. Low-VE datasets, namely Azure VM, Bull, and Hog, have averages below 20\%, with several individual channels near zero, indicating that periodic structure explains little of the observed variability and the residual model must carry most of the generative burden. These tiers recur as an organizing principle throughout the later ablations: template quality, as measured by VE, consistently predicts residual importance, spectral alignment, and the conditions under which SCM mixing helps or degrades. We report VE in percentage points. Columns \emph{Covariate 1}--\emph{Covariate 6} denote the ordered channels used for each dataset; datasets with fewer than six channels use `--` in the remaining entries.

\clearpage
\subsection{t-SNE Ablation}
\label{app:tsne_ablation}

To assess geometric fidelity directly, Figure~\ref{fig:tsne_grid_appendix} visualizes t-SNE projections~\citep{maaten2008visualizing} for all 12 datasets.

\begin{center}
\setlength{\tabcolsep}{4pt}
\renewcommand{\arraystretch}{1.0}
\begin{tabular}{@{}cccc@{}}
\includegraphics[width=0.23\linewidth,height=0.16\textheight,keepaspectratio]{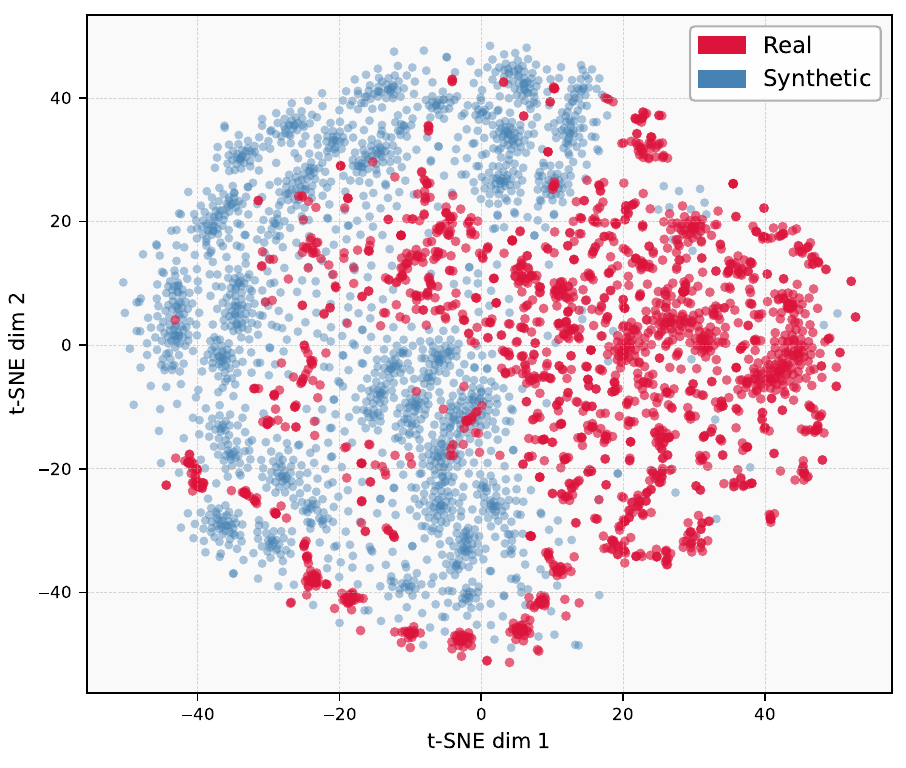} &
\includegraphics[width=0.23\linewidth,height=0.16\textheight,keepaspectratio]{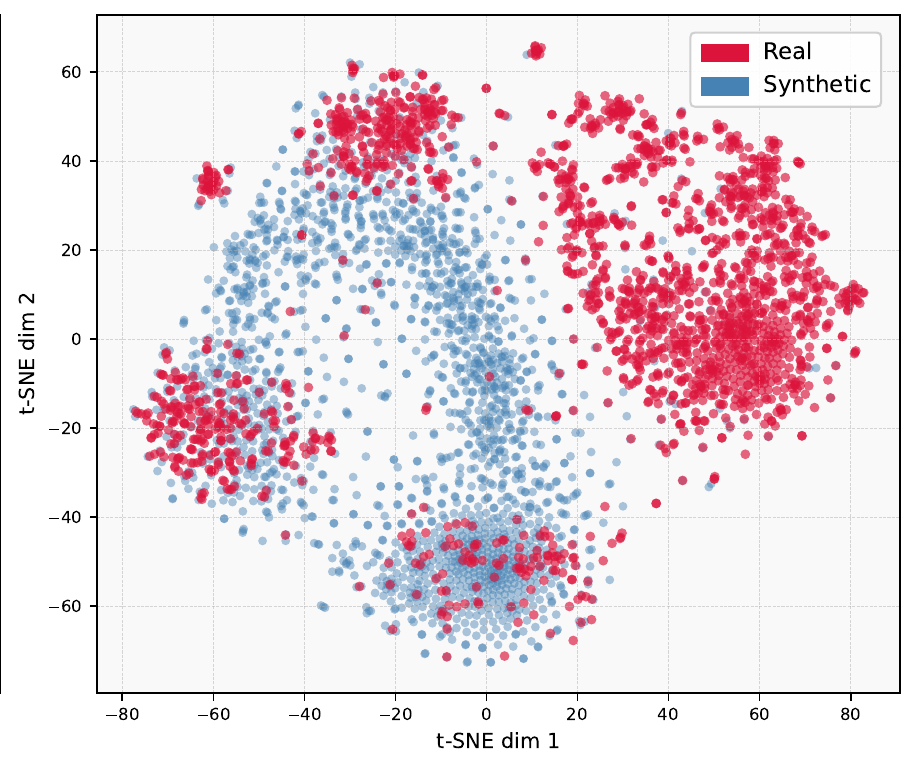} &
\includegraphics[width=0.23\linewidth,height=0.16\textheight,keepaspectratio]{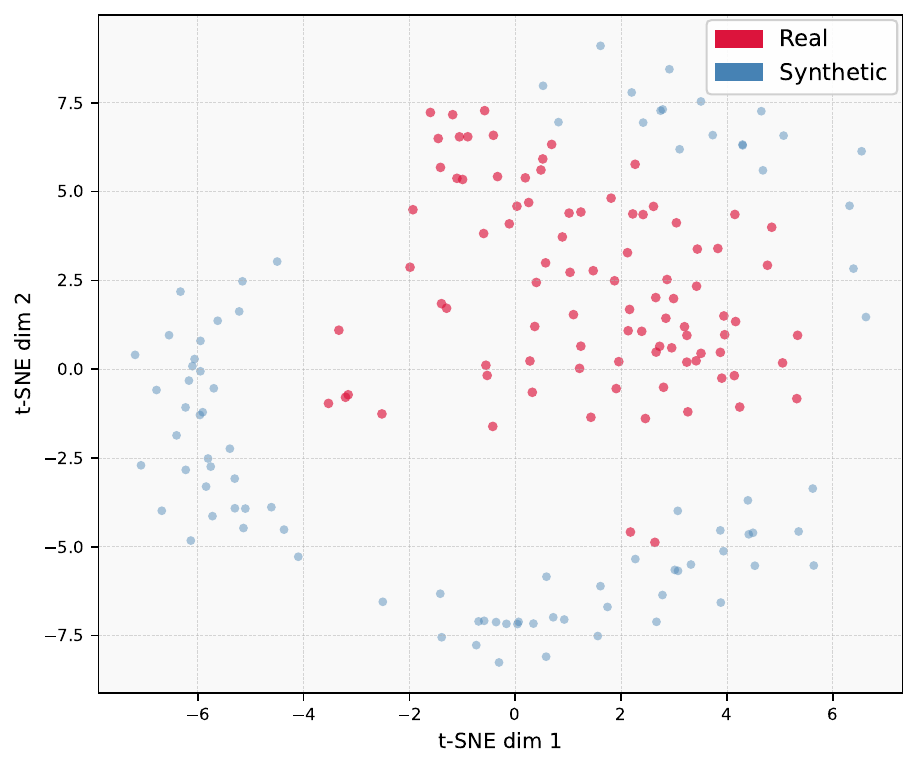} &
\includegraphics[width=0.23\linewidth,height=0.16\textheight,keepaspectratio]{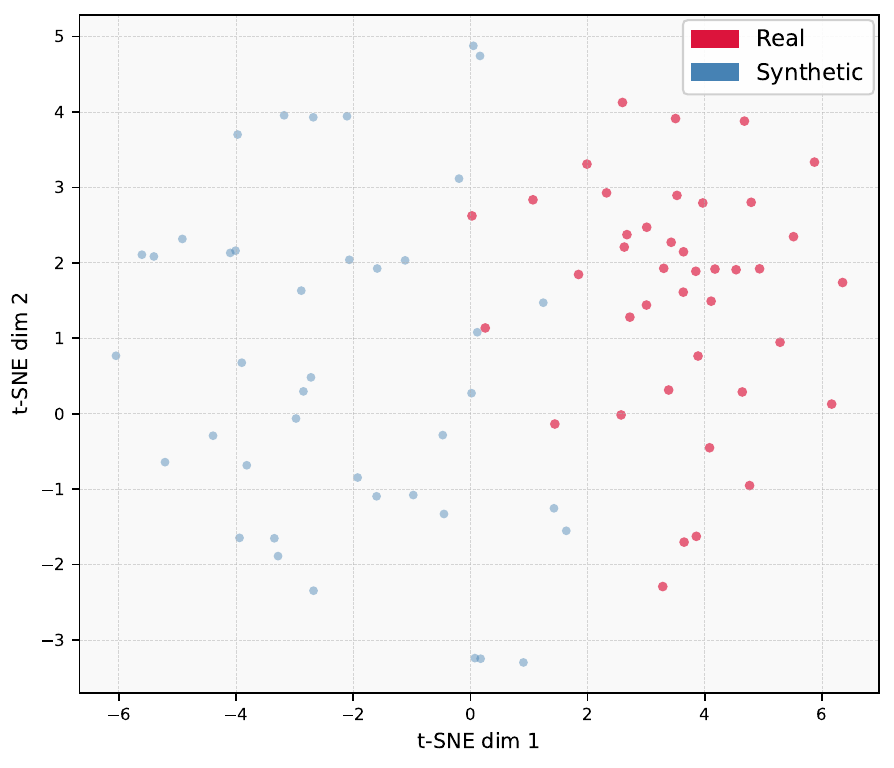} \\
\small Azure VM Traces 2017 & \small Borg Cluster Data 2011 & \small Bear & \small Bull \\
\includegraphics[width=0.23\linewidth,height=0.16\textheight,keepaspectratio]{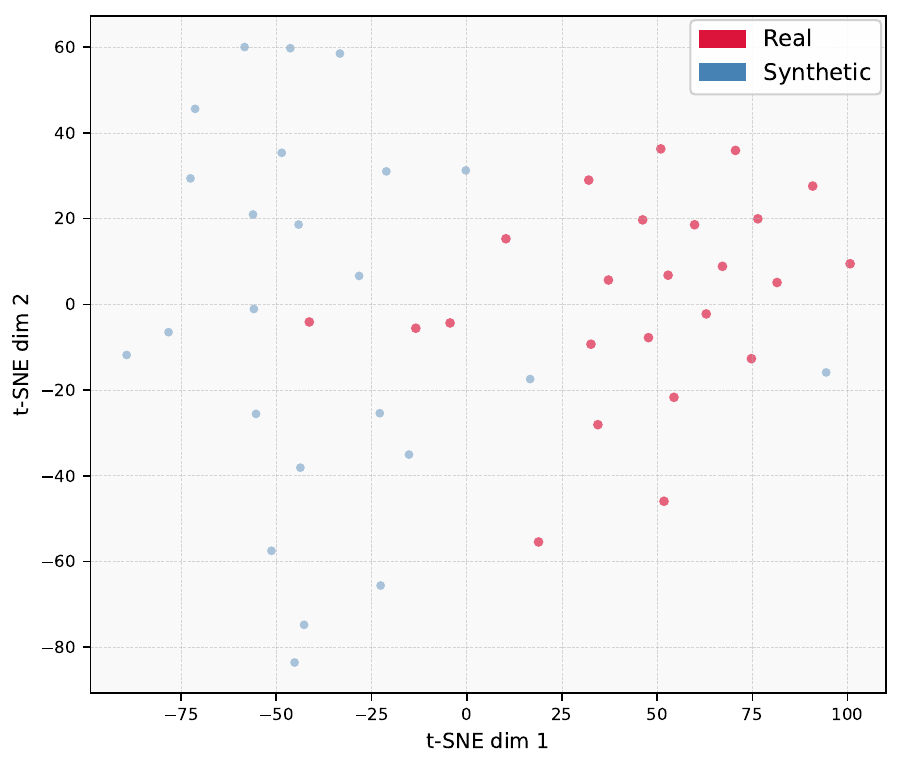} &
\includegraphics[width=0.23\linewidth,height=0.16\textheight,keepaspectratio]{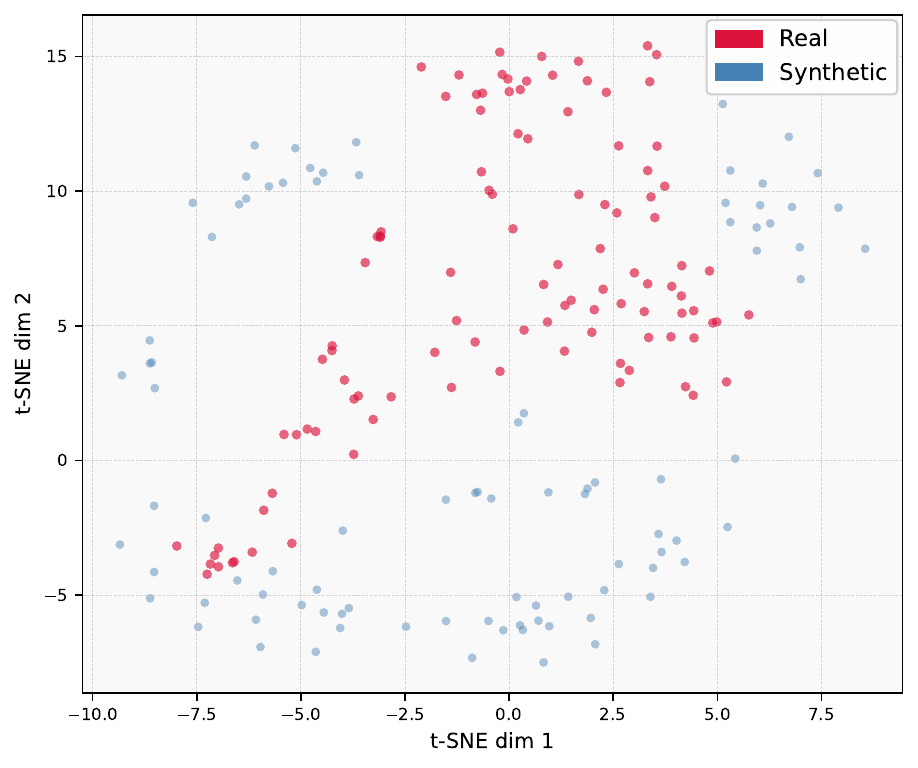} &
\includegraphics[width=0.23\linewidth,height=0.16\textheight,keepaspectratio]{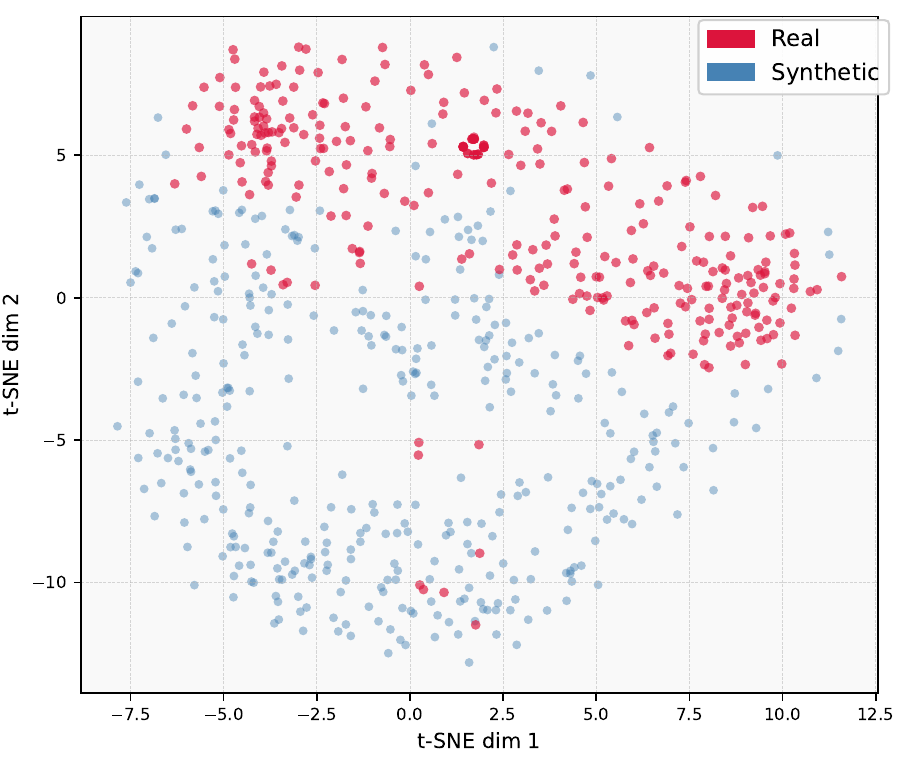} &
\includegraphics[width=0.23\linewidth,height=0.16\textheight,keepaspectratio]{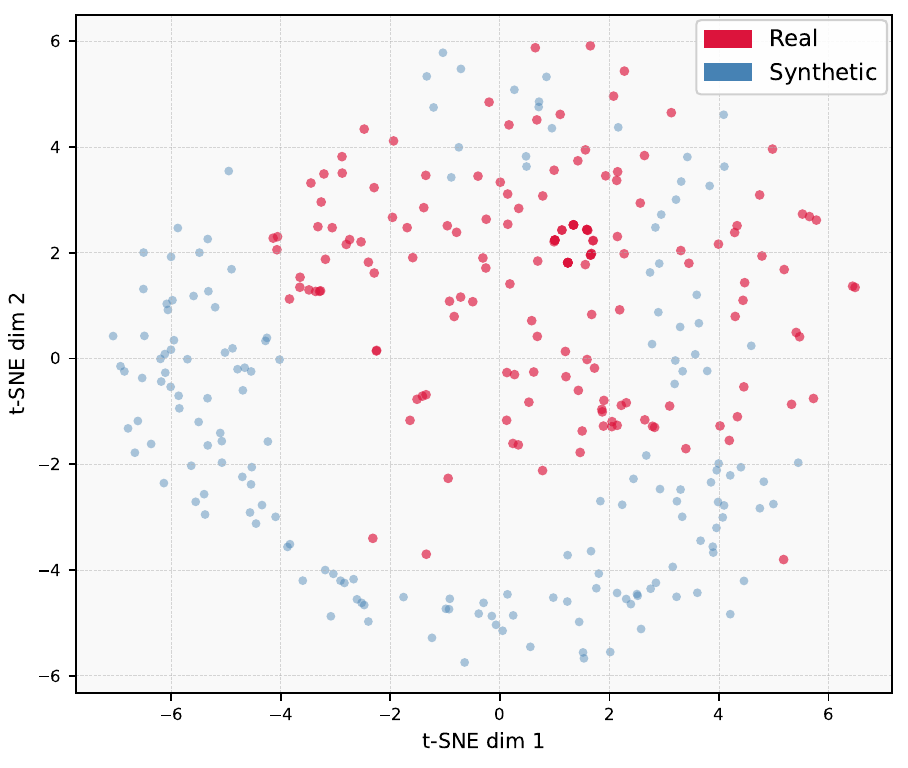} \\
\small Hog & \small Panther & \small PEMS-04 & \small PEMS-08 \\
\includegraphics[width=0.23\linewidth,height=0.16\textheight,keepaspectratio]{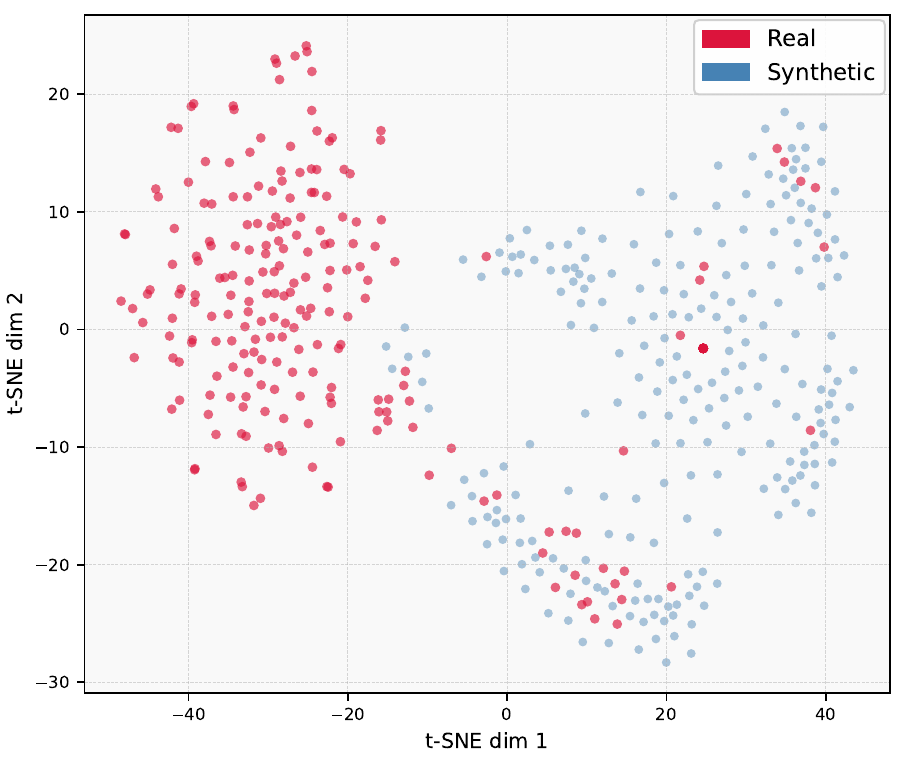} &
\includegraphics[width=0.23\linewidth,height=0.16\textheight,keepaspectratio]{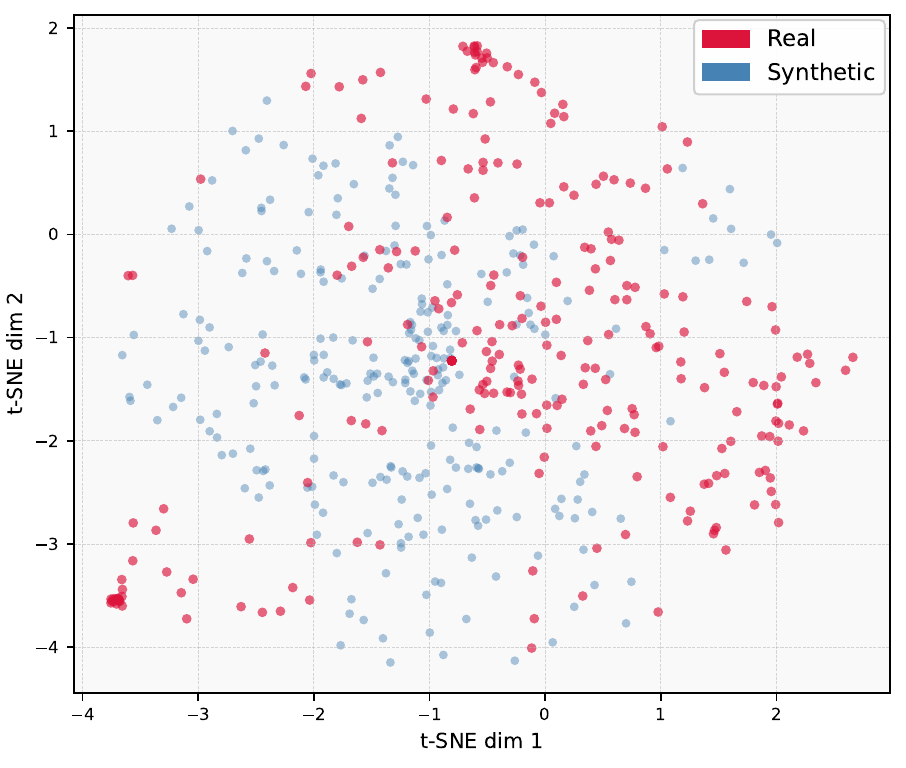} &
\includegraphics[width=0.23\linewidth,height=0.16\textheight,keepaspectratio]{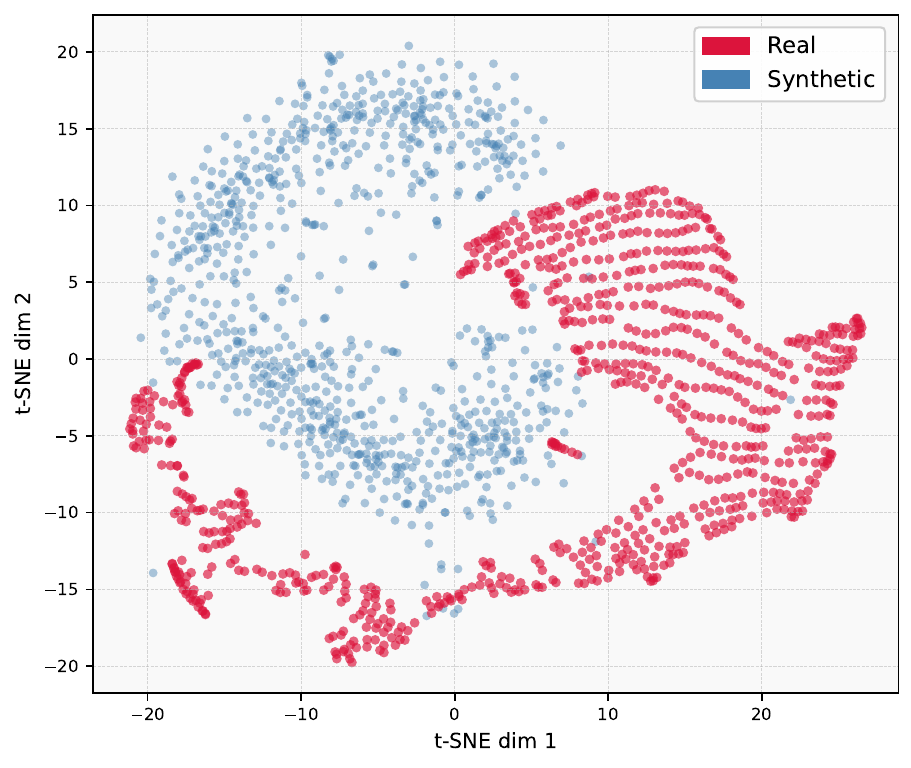} &
\includegraphics[width=0.23\linewidth,height=0.16\textheight,keepaspectratio]{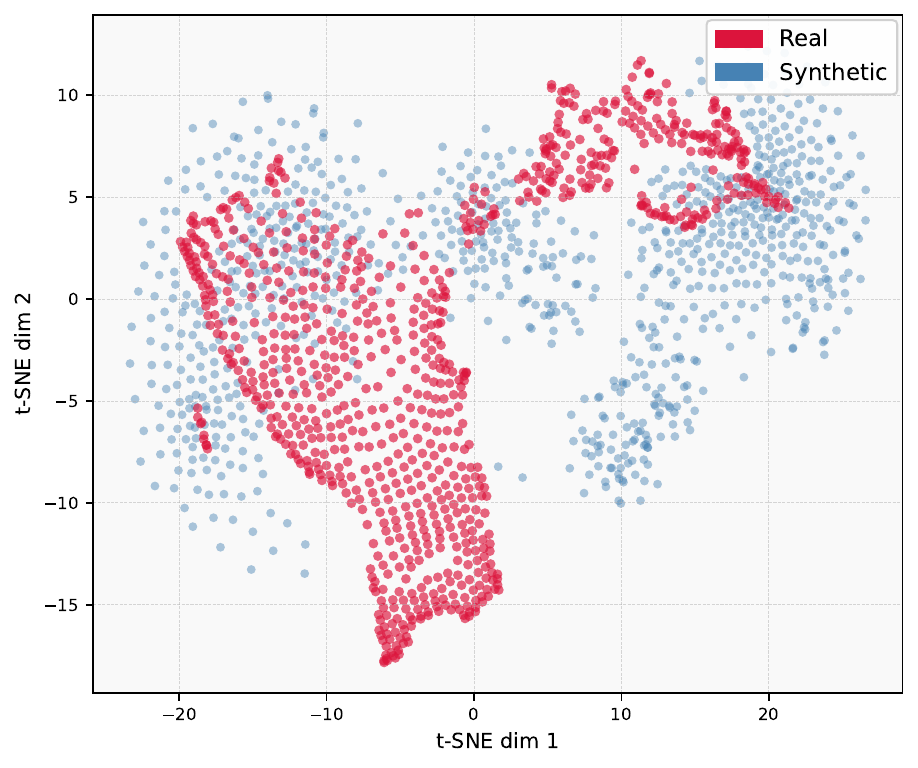} \\
\small Residential PV Power & \small Residential Load Power & \small Subseasonal & \small Subseasonal Precipitation
\end{tabular}
\captionof{figure}{t-SNE projections of real and synthetic samples for twelve representative datasets, illustrating broad geometric alignment together with partial separation in lower-density regions.}
\label{fig:tsne_grid_appendix}
\end{center}

As a qualitative diagnostic, Figure~\ref{fig:tsne_grid_appendix} offers a complementary geometric view of the same phenomenon. Across all twelve datasets, the real and synthetic point clouds remain close in the embedding space without becoming fully superposed, suggesting substantial alignment at the level of coarse support while preserving some separation between the two distributions. The absence of exact overlap is arguably favorable, since a synthetic sample that simply collapsed onto the densest empirical regions of the real data would provide comparatively limited additional coverage.

This pattern is particularly visible in the larger datasets, Azure VM Traces 2017 and Borg Cluster Data 2011. In Azure, the synthetic points extend along the outer envelope and lower-density arcs of the cloud rather than concentrating only in the most populated real regions. In Borg, they occupy several bridging and peripheral regions around the main lobes, again indicating coverage of areas that appear relatively sparse in the observed sample. For the smaller datasets such as Bear, Bull, Hog, and Panther, the two clouds more often come into contact near boundary or transition regions while remaining only partially overlapping overall. Given the reduced sample size and the inherent variability of two-dimensional embeddings, this degree of separation is compatible with the view that the synthetic distribution tracks the same broad geometry while allocating somewhat greater mass to nearby regions that are underrepresented in the real data.

Residential PV Power is the clearest exception, where the separation is more pronounced. The synthetic cloud forms a distinct island anchored around a small subset of real points rather than spreading across the full real manifold. Because the periodic template accounts for 88.6\% of signal variance in this dataset, the t-SNE embedding is driven mostly by residual variation, since the shared periodic structure contributes little to point separation. The over-dispersed residuals produced by the elevated temperature range are therefore more visible here than in other datasets, where residual variation is a smaller fraction of the total signal. This also explains why the forecasting impact remains moderate despite the geometric separation: the template carries most of the predictive signal, leaving relatively little room for the miscalibrated residual to degrade downstream performance. The PV result points to the temperature range as a dataset-specific hyperparameter that would benefit from tuning in domains where the residual component is especially sensitive to sampling stochasticity.

\clearpage
\subsection{Frequency-Domain Ablation}
\label{app:frequency_ablation}

To assess spectral fidelity directly, Figure~\ref{fig:psd_grid_appendix} provides a complementary power-spectral comparison using Welch-style PSD estimates~\citep{welch1967use}.

\begin{center}
\setlength{\tabcolsep}{2pt}
\renewcommand{\arraystretch}{0.95}

{\small\textbf{Residential PV Power}}\\[0.1em]
\begin{tabular}{ccc}
\pspdfthree{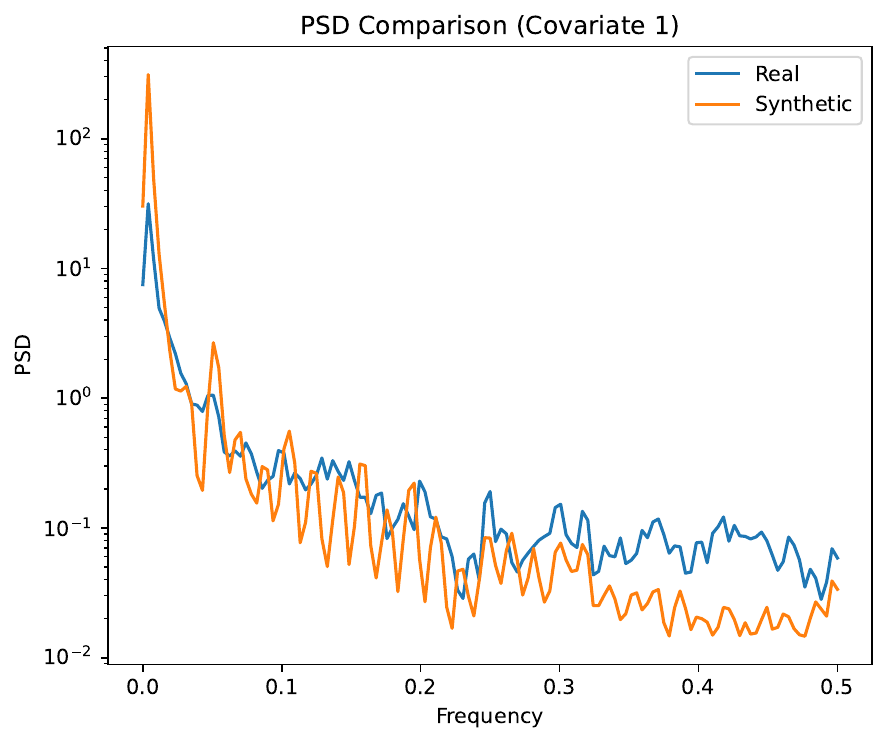}{1} &
\pspdfthree{Diagrams/cropped/residential_pv_power_psdplots-combined.pdf}{2} &
\pspdfthree{Diagrams/cropped/residential_pv_power_psdplots-combined.pdf}{3} \\
\end{tabular}\\[0.45em]

{\small\textbf{Residential Load Power}}\\[0.1em]
\begin{tabular}{ccc}
\pspdfthree{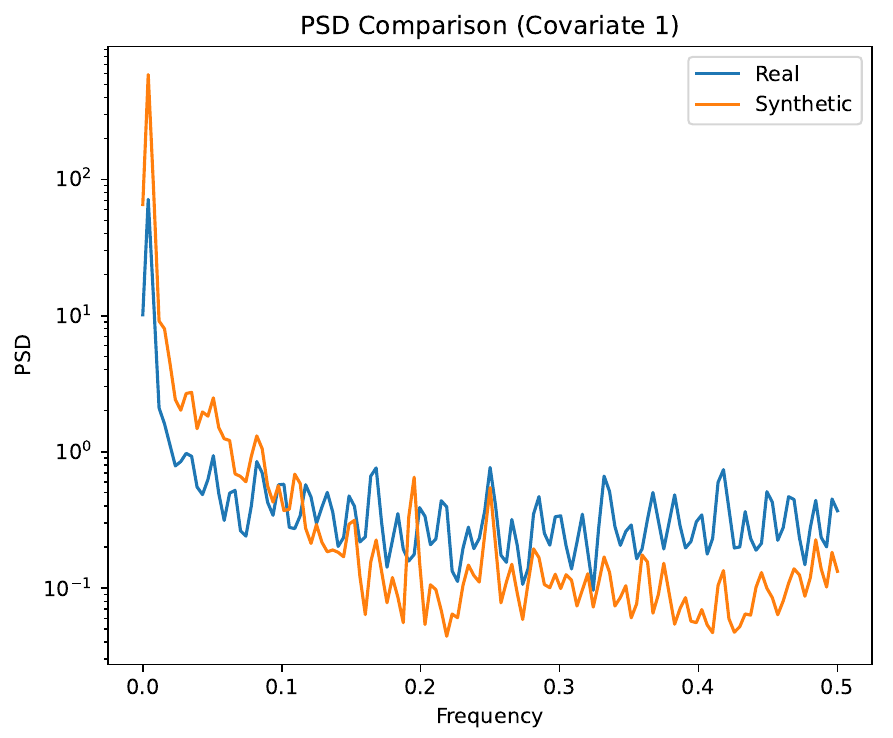}{1} &
\pspdfthree{Diagrams/cropped/residential_load_power_psdplots-combined.pdf}{2} &
\pspdfthree{Diagrams/cropped/residential_load_power_psdplots-combined.pdf}{3} \\
\end{tabular}\\[0.45em]

{\small\textbf{PEMS-04}}\\[0.1em]
\begin{tabular}{ccc}
\pspdfthree{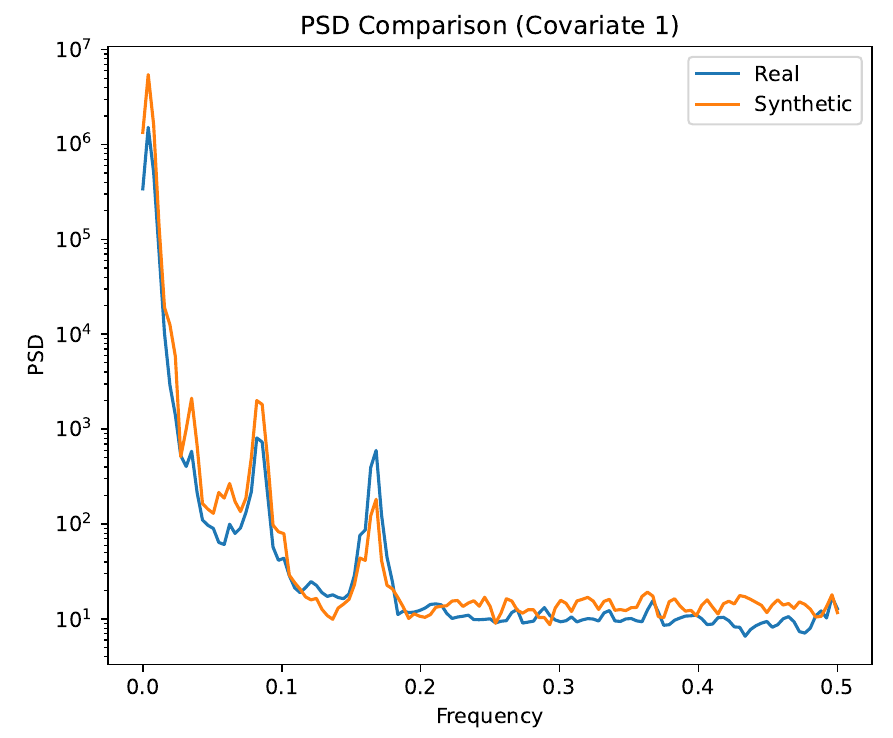}{1} &
\pspdfthree{Diagrams/cropped/pems04_psdplots-combined.pdf}{2} &
\pspdfthree{Diagrams/cropped/pems04_psdplots-combined.pdf}{3} \\
\end{tabular}\\[0.45em]

{\small\textbf{PEMS-08}}\\[0.1em]
\begin{tabular}{ccc}
\pspdfthree{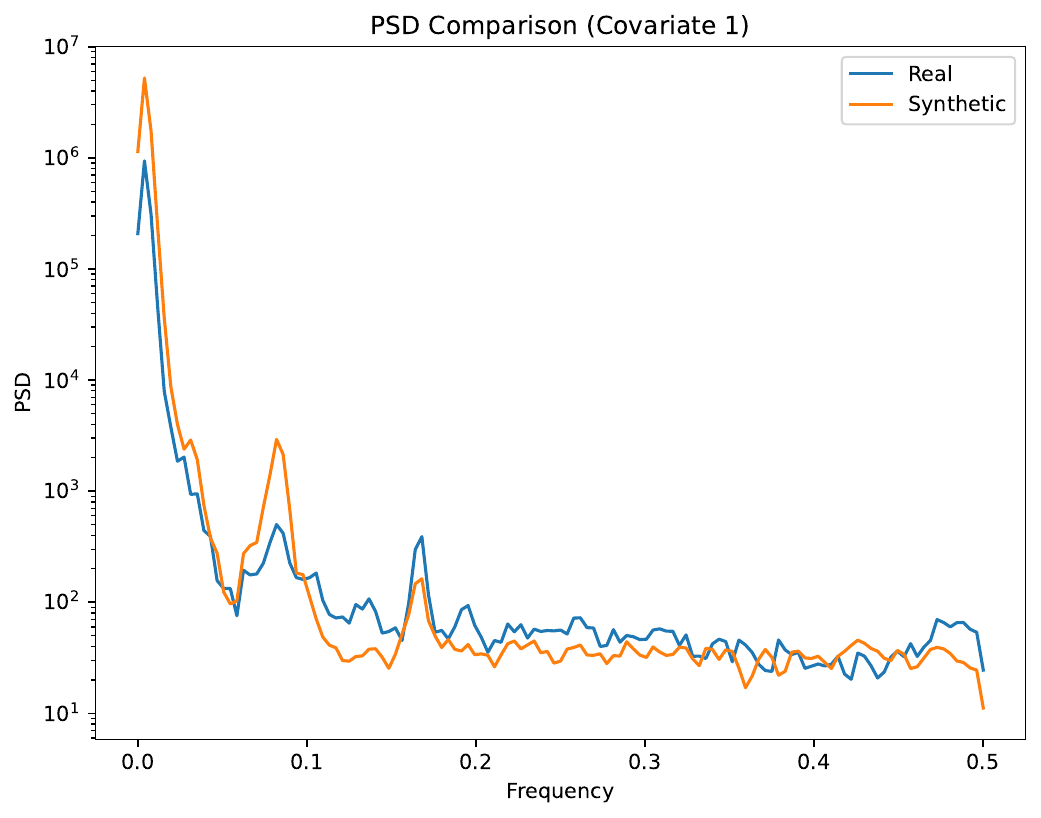}{1} &
\pspdfthree{Diagrams/cropped/pems08_psdplots.pdf}{2} &
\pspdfthree{Diagrams/cropped/pems08_psdplots.pdf}{3} \\
\end{tabular}\\[0.45em]

\newpage
{\small\textbf{Azure VM Traces 2017}}\\[0.1em]
\begin{tabular}{ccc}
\pspdfthree{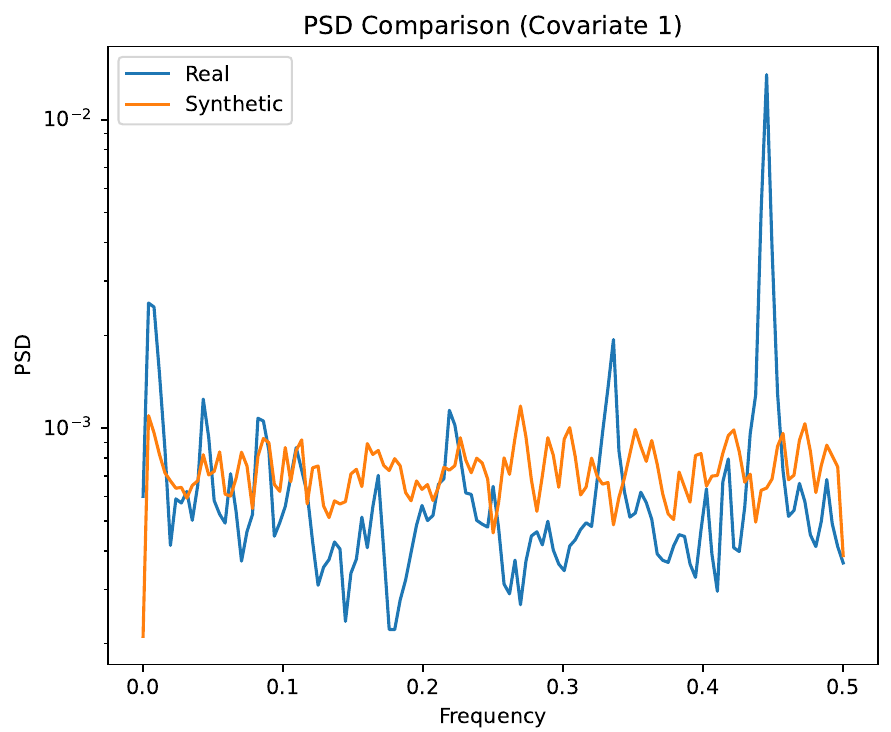}{1} &
\pspdfthree{Diagrams/cropped/azure_vm_traces_2017_psdplots.pdf}{2} &
\pspdfthree{Diagrams/cropped/azure_vm_traces_2017_psdplots.pdf}{3} \\
\end{tabular}\\[0.45em]

{\small\textbf{Borg Cluster Data 2011}}\\[0.1em]
\begin{tabular}{ccc}
\pspdfthree{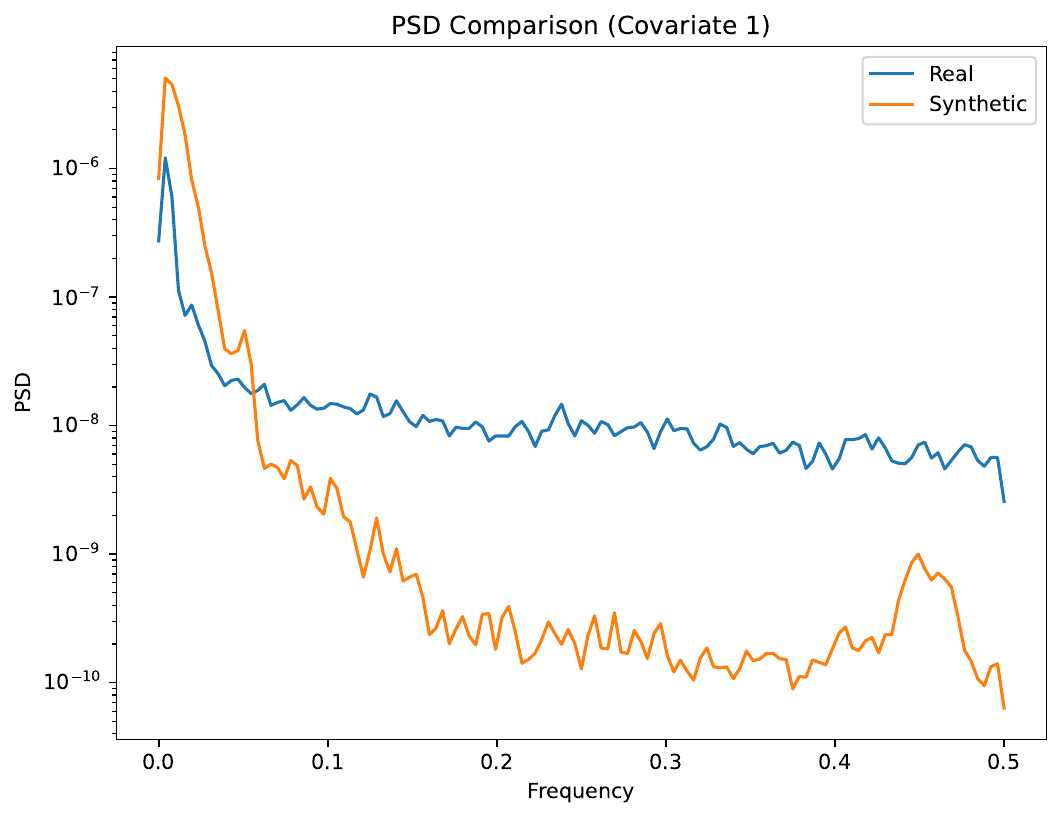}{1} &
\pspdfthree{Diagrams/cropped/borg_cluster_data_2011_psdplots.pdf}{2} &
\pspdfthree{Diagrams/cropped/borg_cluster_data_2011_psdplots.pdf}{3} \\
\pspdfthree{Diagrams/cropped/borg_cluster_data_2011_psdplots.pdf}{4} &
\pspdfthree{Diagrams/cropped/borg_cluster_data_2011_psdplots.pdf}{5} &
\pspdfthree{Diagrams/cropped/borg_cluster_data_2011_psdplots.pdf}{6} \\
\end{tabular}\\[0.45em]

{\small\textbf{Single-covariate datasets}}\\[0.1em]
\begin{tabular}{ccc}
\pspdfthree{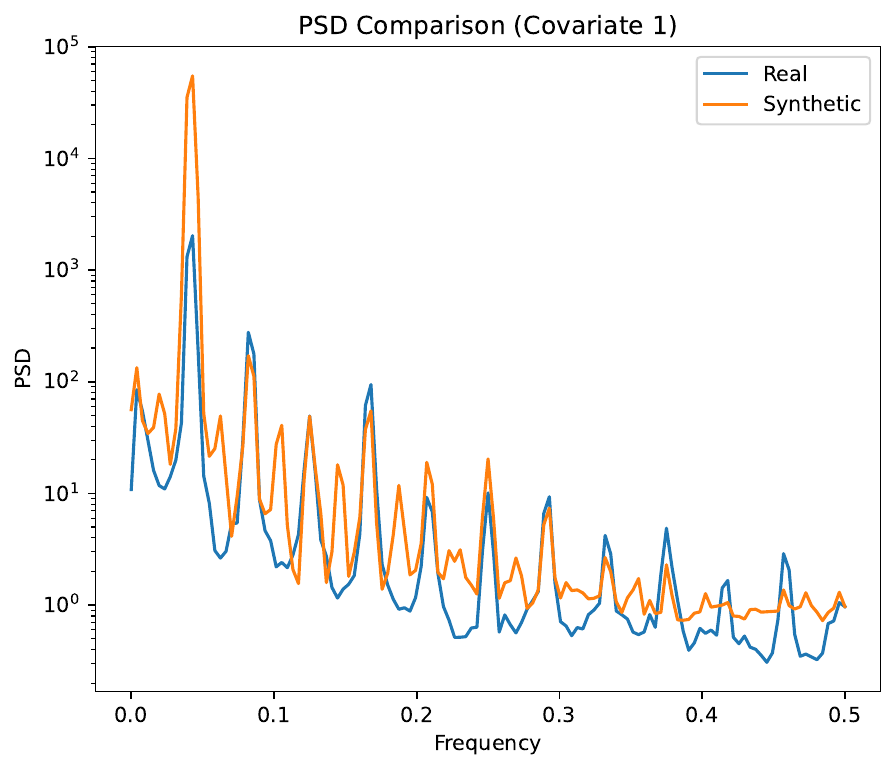}{1} &
\pspdfthree{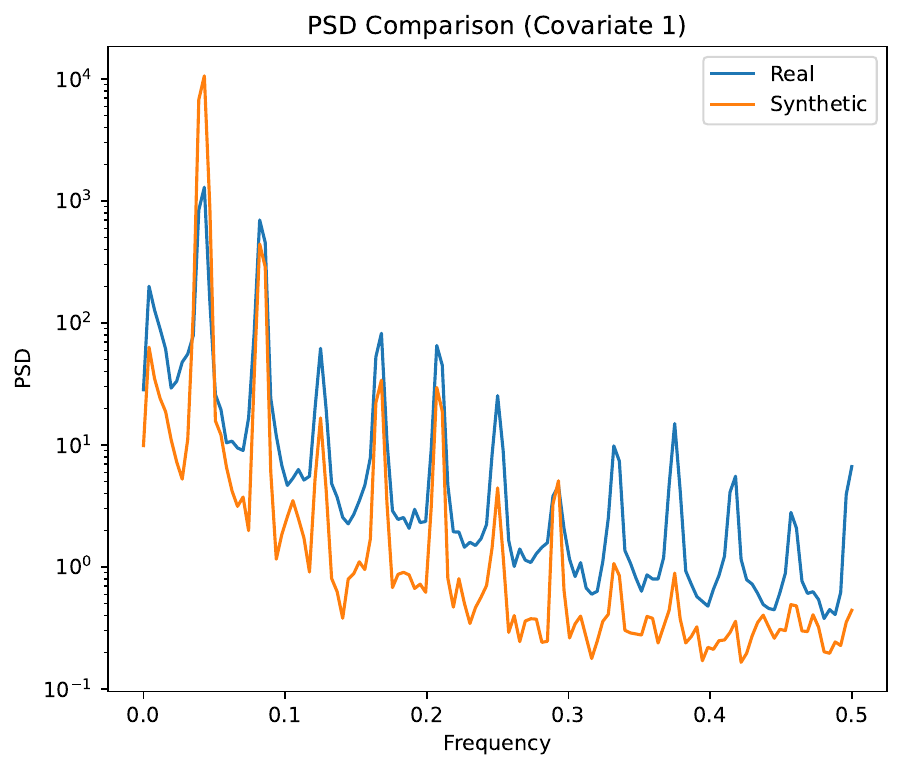}{1} &
\pspdfthree{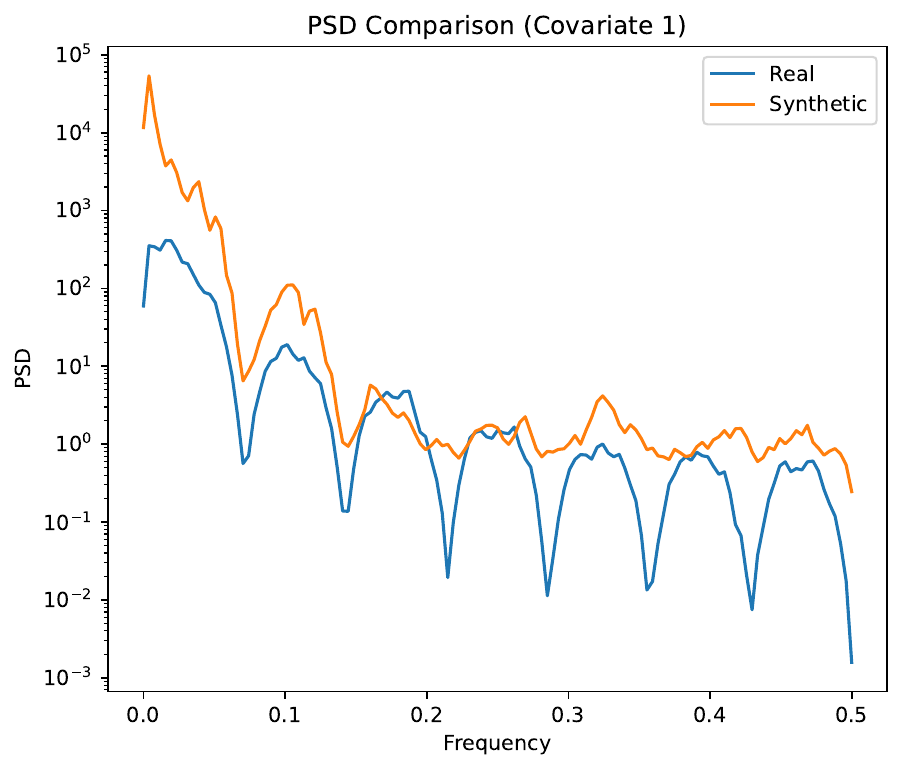}{1} \\
\tiny BDG-2 Bear & \tiny BDG-2 Panther & \parbox[c]{0.30\textwidth}{\centering\tiny Subseasonal Precipitation} \\
\end{tabular}\\[0.45em]

{\small\textbf{BDG-2 Bull}}\\[0.1em]
\begin{tabular}{cccc}
\pspdffour{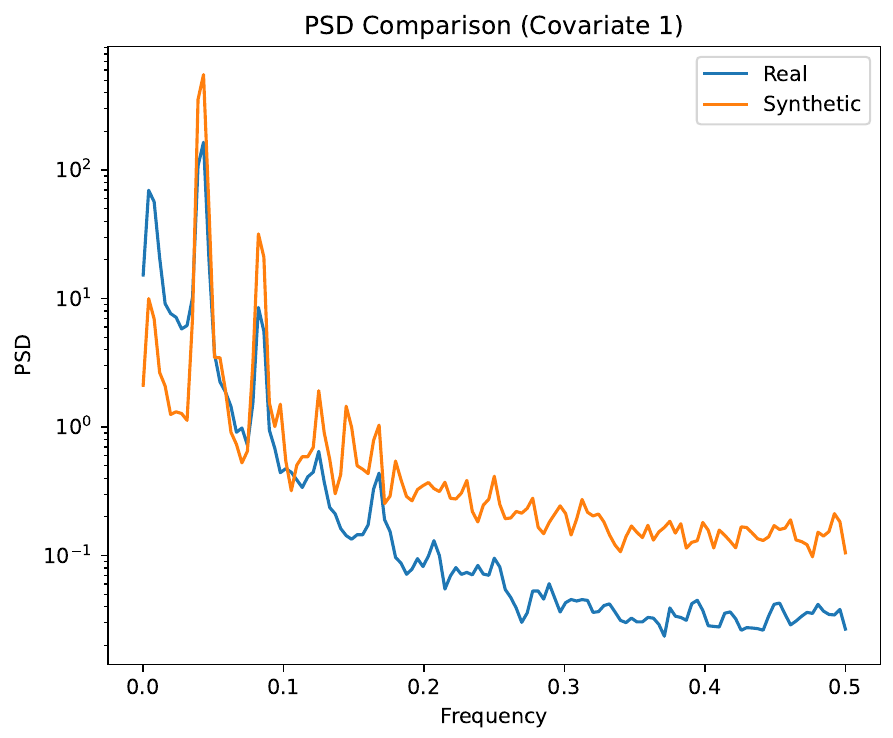}{1} &
\pspdffour{Diagrams/cropped/bull_psdplots-combined.pdf}{2} &
\pspdffour{Diagrams/cropped/bull_psdplots-combined.pdf}{3} &
\pspdffour{Diagrams/cropped/bull_psdplots-combined.pdf}{4} \\
\end{tabular}
\captionof{figure}{Power spectral density (PSD) comparison between real and synthetic time series across representative datasets and covariates. Each subplot shows frequency on the x-axis and spectral power density on the y-axis, with the real series shown in blue and the synthetic series shown in orange. The panels are arranged dataset-by-dataset with at most three subplots per row, except for BDG-2 Bull, which is shown with four panels in one row, and Borg Cluster Data 2011, which spans two rows. Overall, the synthetic series preserve the dominant spectral peaks and low-frequency decay patterns of the real data while allowing controlled deviations across datasets and covariates.}
\label{fig:psd_grid_appendix}
\end{center}

The PSD comparisons in Figure~\ref{fig:psd_grid_appendix} provide a complementary spectral view of the fidelity story told by the forecasting results. Across the included datasets, the synthetic series preserve the dominant spectral structure of the real data well, and the cases where alignment is imperfect can be traced to specific structural properties of the target domain rather than to a general failure of the pipeline.

The clearest successes are the residential energy datasets and the PEMS traffic benchmarks. For Residential PV Power and Residential Load Power (avg.~VE $> 88\%$), the synthetic PSD tracks the real spectral envelope closely across all three covariates in both datasets. The dominant low-frequency decay, harmonic positions, and relative inter-peak power levels are all reproduced well. PEMS-04 and PEMS-08 (avg.~VE $> 80\%$) show the same pattern in a different domain, with tight synthetic-to-real alignment across all three covariates, harmonic peaks at the correct positions, and a spectral floor that stays close to the real one. The fact that this holds for both members of the sibling pair points to genuine structural preservation rather than accidental matching. BDG-2 Bear and BDG-2 Panther, despite more moderate VE values (57.3\% and 39.7\%), also align well. The synthetic PSD reproduces the overall spectral decay and main harmonic positions with only minor excess power at isolated frequencies, indicating that even when the template is not dominant, the DKL residual still learns a spectrally compatible stochastic process rather than introducing spurious structure.

Two cases warrant closer attention. In Borg Cluster Data 2011, the synthetic PSD decays faster than the real one across several covariates, most visibly on covariate 4 (VE 0.7\%). With 10,000 series, averaging to compute $\bar r$ suppresses high-frequency variability aggressively. Individual-series noise cancels in the mean, and the DKL model then learns a smoother residual process than any individual real series exhibits, causing generated residuals to underrepresent the elevated high-frequency floor of the real corpus. Azure VM Traces 2017, despite also containing 10,000 series, does not show the same pattern because its underlying signal is genuinely broadband and aperiodic. Both real and synthetic PSDs are irregular throughout, which is exactly what a faithful reproduction of a noisy process should look like. The over-smoothing effect becomes visible only when the real signal has a sustained spectral floor that the smoothed residual fails to maintain, a condition present in Borg's more structured channels but not in Azure. BDG-2 Bull, despite similarly low per-channel VE, also avoids this issue because its 41-series mean retains considerably more high-frequency structure. The Borg case is therefore specific to very large corpora with genuine mid-to-high-frequency structure, while the dominant low-frequency features remain correctly reproduced.

In Subseasonal Precipitation, the synthetic PSD preserves the seasonal harmonic positions and overall spectral shape, but the real signal has unusually sharp nulls between harmonics that the synthetic does not fully reproduce. This follows from the additive decomposition rather than from a failure of any one component. The real signal is spectrally very pure, behaving almost like a sinusoidal comb with minimal residual energy between harmonics. Once any stochastic residual is added, between-peak power necessarily increases because the residual model cannot enforce destructive interference at specific frequencies. The mismatch is therefore confined to null depth and does not affect the forecasting-relevant low- and mid-frequency envelope.

Taken together, the PSD comparisons confirm that \textsc{ReGeN} reliably preserves the spectral structure most relevant for forecasting across the majority of the included datasets. The cases where alignment is only partial have clear structural explanations, one linked to aggressive averaging in very large corpora and the other to an unusually pure periodic signal where any additive residual raises a spectral floor that would otherwise remain nearly empty. In both cases, the dominant spectral features are still reproduced correctly.

\clearpage
\subsection{Residual Ablation Under TSTR}
\label{app:residual_tstr_ablation}

To isolate the contribution of residual modelling, Table~\ref{tab:tstr_no_residual_ablation} compares the standard TSTR setup against a variant in which synthetic data is generated without sampled residuals, reported only for iTransformer. Removing the DKL residual consistently degrades performance, but the magnitude of that degradation is strongly structured by how much signal variance the periodic template already explains.

\begin{center}
\captionof{table}{Residual ablation under train-on-synthetic, test-on-real transfer (TSTR), reported only for iTransformer. The TSTR baseline values are copied from Table~\ref{tab:compact_all_pairs}; the no-residual columns are provided to show the degradation when residual modelling is removed. Lower MSE and MAE are better.}
\label{tab:tstr_no_residual_ablation}
\small
\setlength{\tabcolsep}{5pt}
\renewcommand{\arraystretch}{1.08}
\resizebox{0.9\linewidth}{!}{%
\begin{tabular}{lcccccc}
\toprule
\multirow{2}{*}{\textbf{Dataset}} & \multicolumn{2}{c}{\textbf{TSTR}} & \multicolumn{2}{c}{\textbf{Without Residual}} & \multicolumn{2}{c}{\textbf{$\Delta$}} \\
\cmidrule(lr){2-3}\cmidrule(lr){4-5}\cmidrule(lr){6-7}
& \textbf{MSE}$\downarrow$ & \textbf{MAE}$\downarrow$ & \textbf{MSE}$\downarrow$ & \textbf{MAE}$\downarrow$ & \textbf{MSE} & \textbf{MAE} \\
\midrule
BDG-2 Bear & 0.41 & 0.43 & 0.45 & 0.46 & +0.04 & +0.03 \\
BDG-2 Panther & 0.36 & 0.38 & 0.39 & 0.42 & +0.03 & +0.04 \\
Azure VM Traces 2017 & 0.90 & 0.45 & 1.02 & 0.49 & +0.12 & +0.04 \\
Borg Cluster Data 2011 & 0.59 & 0.50 & 0.67 & 0.55 & +0.08 & +0.05 \\
PEMS-04 & 0.37 & 0.38 & 0.40 & 0.41 & +0.03 & +0.03 \\
PEMS-08 & 0.30 & 0.29 & 0.31 & 0.30 & +0.01 & +0.01 \\
BDG-2 Bull & 0.37 & 0.40 & 0.52 & 0.59 & +0.15 & +0.19 \\
BDG-2 Hog & 0.50 & 0.48 & 0.58 & 0.58 & +0.08 & +0.10 \\
Subseasonal & 0.40 & 0.43 & 0.41 & 0.43 & +0.01 & +0.00 \\
Subseasonal Precipitation & 1.01 & 0.73 & 1.14 & 0.80 & +0.13 & +0.07 \\
Residential PV Power & 0.26 & 0.20 & 0.26 & 0.19 & +0.00 & -0.01 \\
Residential Load Power & 0.54 & 0.38 & 0.56 & 0.39 & +0.02 & +0.01 \\
\bottomrule
\end{tabular}%
}
\end{center}

The clearest pattern emerges when the residual-removal degradation is related back to the per-dataset variance explained (VE) values in Table~\ref{tab:ve_by_covariate}. Datasets where the phase-aligned template captures little of the original signal variance suffer the largest collapse when residuals are removed: BDG-2 Bull (avg.~VE 5.3\%) degrades by +0.15 MSE and +0.19 MAE, Azure VM Traces 2017 (avg.~VE 16.2\%) by +0.12 MSE and +0.04 MAE, and BDG-2 Hog (avg.~VE 2.8\%) by +0.08 MSE and +0.10 MAE. By contrast, datasets where the template is highly explanatory are largely unaffected: PEMS-08 (avg.~VE 80.3\%) degrades by only +0.01 MSE and +0.01 MAE, Subseasonal (avg.~VE 80.4\%) by +0.01 MSE and +0.00 MAE, Residential PV Power (avg.~VE 88.6\%) shows no meaningful change, and Residential Load Power (avg.~VE 91.2\%) changes by only +0.02 MSE and +0.01 MAE.

To quantify this relationship, we compute the Spearman rank correlation~\citep{spearman1904proof} between per-dataset average VE and the absolute MSE degradation upon residual removal, finding $\rho=-0.8225$. The same relationship holds for MAE degradation, with $\rho=-0.8858$. These coefficients confirm that template quality is a reliable predictor of residual importance: the less structure the template captures, the more load-bearing the residual model becomes.

This has a direct mechanistic interpretation. In high-VE settings such as PEMS, Subseasonal, and the residential PV/load benchmarks, the periodic template dominates the generative signal and the residual model acts mainly as a stochastic correction. In low-VE settings such as Bull, Hog, and Azure VM, the template fails to capture dominant variability---including irregular load spikes, bursty VM utilization, and aperiodic demand shifts---so the DKL residual must carry much more of the generative burden. In these cases, removing it erases the structured variability that makes the synthetic series informative as a training source and leaves the generator with a much more rigid view of the process. The residual model is therefore not uniformly a correction term but, conditionally, a principal generative component whose importance is determined by the degree to which the target domain exhibits stable periodic structure.

\clearpage
\subsection{SCM Mixing Ablation Under TSTR}
\label{app:scm_tstr_ablation}

To isolate the contribution of multivariate structural coupling, Table~\ref{tab:tstr_no_scm_ablation} compares the standard TSTR setup against a variant in which SCM-based mixing is removed during synthetic generation. Since this ablation is meaningful only for multivariate datasets, the univariate BDG-2 Bear, BDG-2 Panther, and Subseasonal Precipitation datasets are omitted. As above, we give the model name in the caption and copy the TSTR baseline values directly from Table~\ref{tab:compact_all_pairs}.

\begin{center}
\captionof{table}{SCM-mixing ablation under train-on-synthetic, test-on-real transfer (TSTR), reported only for iTransformer. The TSTR baseline values are copied from Table~\ref{tab:compact_all_pairs}; the w/o SCM mixing columns are provided to show the degradation when SCM-based mixing is removed. Lower MSE and MAE are better.}
\label{tab:tstr_no_scm_ablation}
\small
\setlength{\tabcolsep}{5pt}
\renewcommand{\arraystretch}{1.08}
\resizebox{0.9\linewidth}{!}{%
\begin{tabular}{lcccccc}
\toprule
\multirow{2}{*}{\textbf{Dataset}} & \multicolumn{2}{c}{\textbf{TSTR}} & \multicolumn{2}{c}{\textbf{Without SCM Mixing}} & \multicolumn{2}{c}{\textbf{$\Delta$}} \\
\cmidrule(lr){2-3}\cmidrule(lr){4-5}\cmidrule(lr){6-7}
& \textbf{MSE}$\downarrow$ & \textbf{MAE}$\downarrow$ & \textbf{MSE}$\downarrow$ & \textbf{MAE}$\downarrow$ & \textbf{MSE} & \textbf{MAE} \\
\midrule
Azure VM Traces 2017 & 0.90 & 0.45 & 0.96 & 0.46 & +0.06 & +0.01 \\
Borg Cluster Data 2011 & 0.59 & 0.50 & 0.62 & 0.54 & +0.03 & +0.04 \\
PEMS-04 & 0.37 & 0.38 & 0.38 & 0.39 & +0.01 & +0.01 \\
PEMS-08 & 0.30 & 0.29 & 0.32 & 0.32 & +0.02 & +0.03 \\
BDG-2 Bull & 0.37 & 0.40 & 0.37 & 0.42 & +0.00 & +0.02 \\
BDG-2 Hog & 0.50 & 0.48 & 0.48 & 0.45 & -0.02 & -0.03 \\
Subseasonal & 0.40 & 0.43 & 0.47 & 0.51 & +0.07 & +0.08 \\
Residential PV Power & 0.26 & 0.20 & 0.30 & 0.29 & +0.04 & +0.09 \\
Residential Load Power & 0.54 & 0.38 & 0.56 & 0.42 & +0.02 & +0.04 \\
\bottomrule
\end{tabular}%
}
\end{center}

The SCM ablation points to a modest but mostly positive effect. Most datasets worsen slightly without SCM mixing, while BDG-2 Hog improves marginally and BDG-2 Bull is nearly unchanged. The clearest explanation is not dimensionality alone but how much evidence is available per candidate edge in the consensus graph. For Hog, the combination of few series and relatively many channels makes edge selection noisy: with $\tau_{\mathrm{freq}}=0.2$, an edge can survive after appearing in only 5 of the 24 series. That is a low bar, so some spurious dependencies can enter the graph and then be propagated across channels during generation. By contrast, Borg is also higher-dimensional but has vastly more series, so each candidate edge is evaluated against much stronger evidence and removing SCM mixing hurts as expected. Bull sits between these cases, with fewer channels and 41 series, which is consistent with its near-zero SCM effect. The PEMS datasets also lie in a better-supported regime and show small but consistent gains from SCM mixing. Overall, the pattern suggests that SCM mixing helps when the consensus graph is well supported by the available data and becomes close to neutral when evidence is limited, with Hog as the clearest failure case. A stricter or adaptive consensus threshold for low-data, higher-dimensional settings is therefore a natural direction for future work.

\end{appendices}

\end{document}